\documentclass[lettersize,journal]{IEEEtran}
\usepackage{amsmath,amsfonts}
\usepackage{algorithmic}
\usepackage{algorithm}
\usepackage{array}
\usepackage[caption=false,font=normalsize,labelfont=sf,textfont=sf]{subfig}
\usepackage{textcomp}
\usepackage{stfloats}
\usepackage{url}
\usepackage{verbatim}
\usepackage{xcolor}
\usepackage{threeparttable}
\usepackage{amsmath}
\usepackage{booktabs}
\usepackage{multirow}
\usepackage{graphicx}
\usepackage{cite}

\newcommand{\eg}{e.\,g.,\ }

\begin{document}



\title{Foundation Model-based Evaluation of Neuropsychiatric Disorders:  A Lifespan-Inclusive, Multi-Modal, and Multi-Lingual Study} 

\author{
Zhongren~Dong$^\#$, Haotian~Guo$^\#$, Weixiang~Xu, \\ Huan Zhao, Zixing~Zhang*

\thanks{The work leading to this research was supported by the National Natural Science Foundation of China under Grant No.~62571184, the Department of Science and Technology of Hunan Province under Grant No.~2025RC6003, the Guangdong Basic and Applied Basic Research Foundation under Grant No.~2024A1515010112, the Changsha Science and Technology Bureau Foundation under Grant No.~kq2402082, the Shenzhen Natural Science Foundation under Grant No.~JCYJ20250604190534043, and the Hunan Provincial Key Research and Development Program No.~2024AQ2041.} 
\thanks{{$^\#$ These authors contributed equally to this work.}}
\thanks{$^*$ Corresponding author: Zixing~Zhang (zixingzhang@hnu.edu.cn). }

\thanks{ Z.~Dong, H.~Guo, W.~Xu, H.~Zhao and Z.~Zhang are with the College of Computer Science and Electronic Engineering, Hunan University, Changsha, 410082, China. \{zrdong, haotianguo, xuweixiang, hzhao, zixingzhang\}@hnu.edu.cn. } 
\thanks{Z.~Zhang is also with the Shenzhen Research Institute, Hunan University, Shenzhen 518000, China, and the Ministry of Education Key Laboratory of Fusion Computing of Supercomputing and Artificial Intelligence, Hunan University, Changsha, 410082, China. } 
}



\maketitle

\begin{abstract}

Neuropsychiatric disorders, such as Alzheimer's disease (AD), depression, and autism spectrum disorder (ASD), are characterized by linguistic and acoustic abnormalities, offering potential biomarkers for early detection. Despite the promise of multi-modal approaches, challenges like multi-lingual generalization and the absence of a unified evaluation framework persist. To address these gaps, we propose FEND (Foundation model-based Evaluation of Neuropsychiatric Disorders), a comprehensive multi-modal framework integrating speech and text modalities for detecting AD, depression, and ASD across the lifespan. Leveraging 13 multi-lingual datasets spanning English, Chinese, Greek, French, and Dutch, we systematically evaluate multi-modal fusion performance. Our results show that multi-modal fusion excels in AD and depression detection but underperforms in ASD due to dataset heterogeneity. We also identify modality imbalance as a prevalent issue, where multi-modal fusion fails to surpass the best mono-modal models. Cross-corpus experiments reveal robust performance in task- and language-consistent scenarios but noticeable degradation in multi-lingual and task-heterogeneous settings. By providing extensive benchmarks and a detailed analysis of performance-influencing factors, FEND advances the field of automated, lifespan-inclusive, and multi-lingual neuropsychiatric disorder assessment. We encourage researchers to adopt the FEND framework for fair comparisons and reproducible research.
 
\end{abstract}


\section{Introduction}

\IEEEPARstart{N}{europsychiatric} disorders, such as Alzheimer's disease (AD), depression, and autism spectrum disorder (ASD), significantly impair cognition, emotion, behavior, and communication \cite{dawson2023international}. AD, the most prevalent neurodegenerative disease, causes memory loss and cognitive decline \cite{scheltens2021alzheimer}, while depression and ASD primarily disrupt emotional regulation and behavior \cite{american2013diagnostic}. The global burden of these disorders is staggering, with a new case of dementia diagnosed every three seconds and cases projected to triple by 2050 \cite{patterson2018world}. In 2019, nearly 1 billion people were affected by mental health conditions, with depression being a significant risk factor for dementia \cite{weidner2023world, reitz2023global}. These disorders account for one-third of the global disease burden, severely reducing patients' quality of life and imposing substantial socioeconomic costs \cite{charlson2016burden}. Early detection and intervention are therefore critical.

AD, depression, and ASD are often accompanied by distinct acoustic and linguistic abnormalities, which serve as potential biomarkers for early detection. For instance, AD patients exhibit slower speech rates and abnormal pitch patterns \cite{kempler2005neurocognitive, hoffmann2010temporal}, while individuals with depression demonstrate narrowed pitch variation and repetitive prosody \cite{cummins2015review}. Similarly, ASD is characterized by atypical pitch modulation and reduced speech fluency \cite{constantino2021social}.
These acoustic changes are often linked to underlying emotional states, as evidenced by research in speech-based emotion recognition \cite{zhang2024re, xu2025gatem2former, gao2024adaptive}.
On the linguistic side, AD patients show declines in language comprehension and lexical diversity, while individuals with depression tend to use more negative emotion words and exhibit reduced syntactic complexity. These abnormalities not only reflect the underlying pathology of these disorders but also provide a critical foundation for the development of multi-modal approaches to early detection and intervention.

Acoustic and linguistic abnormalities, which often manifest early in the disease course, provide a robust basis for early diagnosis and intervention. However, traditional diagnostic methods rely heavily on expert assessments, which are time-consuming, subjective, and difficult to scale \cite{brentani2013autism, smith2013diagnosis}. To address these limitations, computer-aided assessment techniques, leveraging machine learning and signal processing, offer efficient and scalable alternatives for detecting disease-specific patterns~\cite{zhang2024intelligent, dong2024hafformer, cho2022non}.

These techniques not only enhance diagnostic accuracy but also enable large-scale screening, making them a promising solution for early detection of neuropsychiatric disorders.

With the rapid advancement of artificial intelligence and machine learning, multi-modal approaches, such as speech and text analysis, have shown considerable promise in the study of neuropsychiatric disorders \cite{wang2021modular, rohanian2021alzheimer, yin2019multi, ray2019multi}. These approaches leverage acoustic foundational models (\eg WavLM~\cite{chen2022wavlm}) to extract speech representations and linguistic foundational models (\eg E5~\cite{wang2022text}) to derive semantic representations~\cite{mohamed2022self}. Pre-trained on large-scale datasets, these models capture rich acoustic and linguistic features, which can be transferred to downstream tasks through techniques such as feature concatenation, attention mechanisms, or multimodal fusion. 
However, despite promising advancements, the field faces critical limitations that hinder practical applicability and generalizability. Research is often fragmented, with studies confined to single datasets, specific languages (predominantly English), or proprietary evaluation metrics \cite{zhou2022tamfn, zhou2023caiinet, han2023spatial, yang2024mmpf}. This creates a landscape where direct, fair comparison of methods is nearly impossible, obscuring the true state of progress and making it difficult for researchers to build upon prior work. The lack of a unified, multi-dimensional benchmark has become a major bottleneck, preventing a systematic understanding of how modern foundation models perform across diverse diseases, languages, and age groups. This gap highlights an urgent need for a robust, generalizable, and standardized framework to address the challenges of cross-dataset, multi-modal, and multi-linguistic analysis in neuropsychiatric disorder assessment.


Crucially, while neuropsychiatric disorders like AD, Depression, and ASD are distinct in their clinical nature, they often share overlapping behavioral dimensions observable through speech and language, such as altered prosody and reduced linguistic complexity~\cite{de2020artificial, trayvick2024speech}. This symptomatic overlap and high comorbidity, particularly between depression and both AD~\cite{mueller2018connected} and ASD~\cite{chandrasekhar2015challenges}, pose significant challenges for differential diagnosis. Consequently, a growing body of work is exploring computational tools to distinguish these conditions~\cite{ehghaghi2022data}. A unified evaluation framework, therefore, offers unique advantages over disorder-specific approaches. It allows for a systematic assessment of a foundation model's ability to capture these cross-cutting biomarkers and its generalizability across different but related medical tasks. Furthermore, it fosters methodological cross-pollination, where insights from one disorder can inform research on another, and provides a holistic view of the current technological landscape, revealing shared challenges and opportunities. It is with this motivation for a broad, comparative, and generalizable assessment that we propose our framework.

To address this critical gap, we introduce FEND (Foundation model-based Evaluation of Neuropsychiatric Disorders), a comprehensive framework designed to systematically evaluate acoustic and linguistic foundation models for neuropsychiatric disorder detection. FEND provides a unified benchmark across AD, depression, and ASD, spanning multiple languages and the full human lifespan. By establishing a standardized methodology for preprocessing, evaluation, and analysis, FEND facilitates fair comparisons and reproducible research. Our work makes the following principal contributions:

\begin{itemize}
    \item We introduce FEND, the first comprehensive framework, to our knowledge, for the systematic, multi-dimensional (lifespan, modality, language) evaluation of foundation models in neuropsychiatric disorder detection. It addresses the critical need for standardized benchmarking by providing a unified platform for fair comparison.
    
    \item We present one of the most extensive benchmarks to date, assessing numerous state-of-the-art foundation models on 13 public datasets. This benchmark provides invaluable empirical data on model performance under consistent conditions.
    
    \item Beyond reporting metrics, we provide a deep analysis of the results, identifying crucial factors like modality imbalance and cross-lingual generalization challenges. These insights offer concrete guidance for future model selection and methodology design.
    
    \item By establishing this benchmark and committing to releasing our code, we provide a "ruler" for the community, aiming to enhance the reproducibility and transparency of research in the automated assessment of neuropsychiatric disorders.
\end{itemize}

The remainder of this article is structured as follows. Section~\ref{sec:related_work} reviews related work on mono-modal and multi-modal approaches for neuropsychiatric disorder detection. Section~\ref{sec:benchmark} introduces the FEND framework, detailing its design and methodology. Section~\ref{sec:experimental} describes the experimental setup, including datasets, evaluation metrics, and implementation details. Section~\ref{sec:results} presents a comprehensive analysis of the experimental results, covering mono-modal benchmarks, multi-modal fusion, and cross-corpus generalization. Finally, Section~\ref{sec:conclusion} concludes the article by summarizing key findings and outlining future research directions.

\vspace{-.3cm}
\section{RELATED WORK}\label{sec:related_work}
This section reviews existing research on neuropsychiatric disorder detection. We first introduce the latest advancements of mono-modal and multi-modal methods as well as as their limitations, and finally discuss benchmarks related to neuropsychiatric disorders.

\vspace{-.3cm}
\subsection{Mono-Modal Approaches for Neuropsychiatric Disorder Detection}
\subsubsection{AD}
Mono-modal AD recognition tasks primarily focus on speech and text modalities. In the speech modality, studies either use traditional features (\eg MFCC) with classical classifiers \cite{chen2023cross, balagopalan2021comparing} or leverage self-supervised learning (SSL) models (\eg WavLM) to extract deep audio features \cite{pan2021using, li2023leveraging}. In the text modality, pre-trained language models (\eg BERT) are widely used to capture cognitive impairments \cite{wang2022exploring, syed2021tackling}. However, these methods are limited by their reliance on mono-modal information and lack of standardized evaluation protocols, restricting their generalizability \cite{mei2023ustc}.

\subsubsection{Depression}
Depression detection in the speech modality has evolved from traditional features (\eg MFCC and prosodic features) combined with classical classifiers (\eg SVM) or deep learning models (\eg CNN) \cite{zhao2019automatic, cummins2011investigation, aharonson2020automated} to SSL models, which capture deep-level speech patterns \cite{huang2024depression, chen2023speechformer++, zhang2024paralbench}. In the text modality, Transformer-based models are widely used to analyze social media data \cite{kour2022hybrid, jiang2020detection}. Despite these advancements, challenges remain in multi-modal fusion and multi-lingual generalization, as most studies are based on specific languages or limited datasets \cite{rejaibi2022mfcc, stasak2022breaking}.

\subsubsection{ASD}
ASD detection has primarily focused on the speech modality, with limited exploration of text and multi-modal approaches. Early research relied on manual feature extraction methods (\eg eGeMAPS) combined with classical machine learning classifiers \cite{beccaria2022extraction, baird2017automatic, mohanta2022analysis}. The ASD sub-task of the ComParE 2013 \cite{schuller2013interspeech} substantially advanced the development of automatic detection technologies based on children's speech recordings. However, compared to AD and depression, ASD research lags behind in terms of modeling and methodological innovation. Current studies predominantly rely on traditional acoustic features and classical machine learning methods, with limited exploration of SSL techniques, foundational models, and multi-modal fusion. Additionally, due to the unique nature of ASD datasets, research on the text modality remains relatively scarce.

In summary, mono-modal detection methods for AD, depression, and ASD face common limitations:
(1) Insufficient utilization of multi-modal information, restricting the model's ability to capture the complexity of these disorders.
(2) Lack of unified standards in datasets, feature processing, and evaluation methods, hindering comparability and generalizability.
(3) Lag in ASD research, particularly application of SSL, foundational models, and multi-modal fusion.

\vspace{-.2cm}
\subsection{Multi-Modal Approaches for Neuropsychiatric Disorder Detection}

\subsubsection{AD}
Multi-modal AD recognition tasks typically integrate speech and text modalities. Early studies leveraged deep learning models for feature extraction and fusion, such as modular multi-modal attention networks \cite{wang2021modular}. More advanced methods based on attention mechanisms and Transformers have also shown strong potential in improving diagnostic performance \cite{ilias2023context, ilias2023detecting}. More recently, sophisticated approaches have emerged to tackle specific fusion challenges. For example, Ortiz-Perez et al. \cite{ortiz2025cognialign} introduced a word-level alignment strategy using transcription timestamps, enabling a finer-grained fusion with gated cross-attention. Concurrently, Dong et al. \cite{dong2025modality} proposed a dynamic knowledge distillation method to explicitly address the modality imbalance problem by adaptively reinforcing the weaker modality. However, most studies are based on small-scale datasets (\eg ADReSS~\cite{luz2021alzheimer}) and focus on English-speaking populations, leaving multi-linguistic and cross-cultural generalizability unaddressed. Additionally, there is room for optimization in fusion strategies, such as early versus late fusion.

\begin{table*}[!t]
  \centering
  \caption{Summary of Benchmark Studies for Neuropsychiatric Disorder Detection}
  \label{tab:benchmark}
  \begin{threeparttable}
      \begin{tabular}{cccccc}
        \toprule
         Benchmark  &Year   &Task  &Modality  &Language \\
        \midrule
        PROB-SP \cite{de2024probing}     &2024 &Depression  &Speech &French, Italy \\
        SSE-Dep \cite{dumpala2024self}     &2024 &Depression  &Speech   &English \\
        SSM-Sch \cite{premananth2024self}     &2024 &Schizophrenia  &Speech, Video   &English \\
        \midrule 
        FEND      &2024 &AD, Depression, ASD  &Speech, Text &English, Chinese, Greek, French and Dutch \\
      \bottomrule
    \end{tabular}
\end{threeparttable}
\vspace{-.3cm}
\end{table*}

\subsubsection{Depression}
Multi-modal depression detection integrates speech, text, and video modalities. SSL models (\eg BERT, Wav2vec 2.0) have markedly improved feature extraction by capturing latent patterns in language and speech \cite{ray2019multi}. Fusion methods have evolved from simple feature concatenation to complex models based on deep learning and self-attention mechanisms. For example, Ma et al. \cite{yin2019multi} proposed a multi-modal hierarchical recurrent neural network, while the DepMSTAT \cite{tao2024depmstat} leverages spatiotemporal attention mechanisms. Despite these advancements, challenges remain in dataset scale, evaluation standardization, and cross-cultural generalization. The application of foundation models in depression detection is still in its early stages, requiring further research to validate their practicality.

\subsubsection{ASD}
Multi-modal research on ASD is notably scarce, with most studies focusing on mono-modality. The unique challenges in constructing ASD datasets have limited the exploration of multi-modal approaches, particularly the integration of speech and text modalities. This gap highlights the need for further research in foundational models and multi-modal fusion for ASD detection.

In summary, multi-modal methods for AD, depression, and ASD detection have made substantial progress, particularly in feature extraction and fusion based on deep learning. However, existing research shares several common limitations:
(1) Limited dataset scale, constraining model generalization and robustness.
(2) Lack of unified standards in evaluation metrics, data preprocessing, and model training.
(3) Insufficient cross-cultural generalization, with most studies focusing on English-speaking populations.
(4) A notable gap in ASD research, particularly in multi-modal approaches combining speech and text modalities.

\vspace{-.2cm}
\subsection{Benchmark for Neuropsychiatric Disorder Detection}
Foundation Models provide powerful pre-trained feature extraction capabilities for research on AD and mental health. Compared to traditional features, these models capture richer contextual semantic information, significantly improving performance across multiple tasks. Benchmark studies, through standardized datasets and evaluation metrics, offer a unified comparison framework, addressing the issue of non-comparability caused by differences in data sources and preprocessing. This facilitates the evaluation of foundation models' performance across tasks, languages, and modalities, including cross-cultural adaptability and multi-modal generalization.

Despite the potential of foundation models, notable gaps remain in multi-modal and multi-linguistic research. For instance, studies have explored the transferability of pre-trained speech models in mental health detection, analyzing the impact of feature levels, audio segment lengths, and pooling strategies on task performance, achieving state-of-the-art (SOTA) results in depression detection \cite{de2024probing}. Additionally, research based on multitask learning and SSL speech models has demonstrated the performance improvements of SSL embeddings in predicting individual symptoms of depression \cite{dumpala2024self}. In the multi-modal direction, speech representation learning systems based on VQ-VAE have been proposed to predict symptom categories and severity in schizophrenia \cite{premananth2024self}. Meanwhile, Benchmark frameworks for general paralinguistic tasks, such as emotion recognition and age prediction, are also evolving~\cite{zhang2024paralbench}. 
Thus, Table~\ref{tab:benchmark} summarizes related studies in the literature. 

However, most studies are limited to the speech modality, with limited exploration of multi-modal fusion and multi-linguistic adaptability. Addressing these gaps is crucial for advancing the field and enabling robust, scalable, and culturally adaptable multi-modal approaches for neuropsychiatric disorder assessment.


\begin{figure}[!t]
  \centering
  \includegraphics[width=\linewidth]{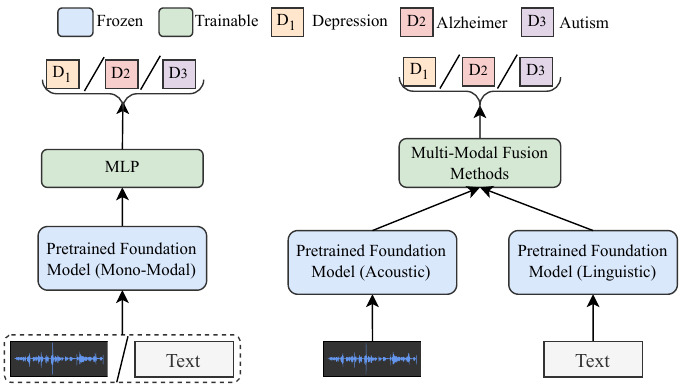}
  \caption{Overview of the FEND Framework for Neuropsychiatric Disorder Detection. The FEND framework includes two main processes: Mono-modal and multi-modal detection. In the mono-modal process, speech or text is fed into a foundation model to extract representations, followed by a disease classification using an MLP. In the mono-modal process, speech and text are separately processed by foundation models to extract representations, which are then fused using classical multi-modal methods for a final disease prediction.}
  \label{framework}
  \vspace{-.4cm}
\end{figure}

\section{Benchmark}\label{sec:benchmark}
This section introduces the FEND framework, a multi-modal neuropsychiatric disorder evaluation framework based on foundation models. The framework is divided into three parts: an overview of its design, the employed acoustic and linguistic foundational models, and the characteristics of the detection tasks for AD, depression, and ASD, along with the datasets and unified evaluation metrics.

\vspace{-.3cm}
\subsection{Framework}
The FEND framework systematically evaluates the performance of foundation models in neuropsychiatric disorder detection tasks through two core processes: mono-modal analysis and multi-modal fusion. By leveraging pre-trained foundation models and carefully designed feature processing modules, FEND addresses key challenges in mono-modal feature representation and multi-modal fusion. As shown in Fig.~\ref{framework}, the framework consists of two main modules: the mono-modal analysis module and the multi-modal fusion module.

\subsubsection{Mono-Modal Analysis Module}
The mono-modal analysis module evaluates the ability of foundation models to capture disease features in a single modality (speech or text). We employ a simple three-layer MLP as the classifier, consisting of an input layer, a hidden layer (256 neurons with ReLU activation), and an output layer. This design ensures that the effectiveness of foundation models is assessed without the complexity of advanced classifiers.

For a given speech modality $A$ and text modality $T$, the feature extraction and modeling process involves two steps:

Step 1: Foundation Model Feature Extraction. Speech or text is fed into a pre-trained foundation model $F$ (\eg WavLM for speech and E5 for text) to extract representations $\mathbf{e}_{m}$:
\begin{equation}
\mathbf{e}_{m} = F_{m}\left ( D; \mathbf{\theta}_{frozen}  \right ), m, D\in \left \{ speech,text \right \},
\end{equation}
where $F_{m}$ represents the pre-trained foundation model, $\mathbf{\theta}_{frozen}$ denotes the fixed pre-trained parameters, and $\mathbf{e}_{m}$ is the extracted modality feature.

Step 2: The extracted representations $\mathbf{e}_{m}$ are then fed into the MLP to obtain the final mono-modal prediction result $y_{mono}$:
\begin{equation}
y_{mono} = MLP\left (\mathbf{e}_{m};\mathbf{\theta} _{trainable}  \right ),
\end{equation}
where $\mathbf{\theta} _{trainable}$ represents the trainable parameters of the MLP, and $y_{mono}$ denotes the predicted output for the mono-modal classification task.




\subsubsection{Multi-Modal Fusion Module}
The multi-modal fusion module integrates speech and text representations to enhance disease detection performance. To provide a comprehensive benchmark, we employ a diverse set of classical fusion algorithms, which can be broadly categorized based on the granularity of fusion. 

\textbf{Sequence-Level Fusion Methods} operate at the frame level to capture fine-grained temporal interactions between modalities. These include MFN~\cite{zadeh2018memory}, which utilizes a gated memory system to model long-term dependencies; Graph-MFN~\cite{zadeh2018multimodal}, an extension that models inter-modal relationships with a dynamic graph; and MulT~\cite{tsai2019multimodal}, which leverages cross-modal attention to effectively fuse unaligned sequences. Concurrently, methods like MFM~\cite{tsai2018learning} focus on disentangling modality-specific and shared representation information.

\textbf{Utterance-Level Fusion Methods} operate on holistic feature vectors representing entire utterances to capture high-level interactions. These include TFN~\cite{zadeh2017tensor}, which explicitly models high-order inter-modal correlations through a tensor outer product; LMF~\cite{liu2018efficient}, which provides an efficient low-rank approximation to TFN; MMIM~\cite{han2021improving}, an information-theoretic approach that maximizes the mutual information between modalities to ensure comprehensive fusion; and a general Attention mechanism~\cite{vaswani_attention} that dynamically weights feature importance. This diverse selection allows us to systematically evaluate a wide range of fusion strategies and their effectiveness at different representational granularities.

Step 1: Foundation Model Feature Extraction. Speech and text are separately passed through their respective pre-trained foundation models $F_{speech}$ and $F_{text}$ to extract representations, resulting in speech feature $\mathbf{e}_{s}$ and text feature $\mathbf{e}_{t}$:
\begin{equation}
\mathbf{e}_{s} = F_{speech}\left ( A;\mathbf{\theta} _{frozen}  \right ), \mathbf{e}_{t} = F_{text}\left ( T;\mathbf{\theta} _{frozen}  \right ).
\end{equation}

Step 2: Multi-Modal Fusion Modeling. The extracted features $\mathbf{e}_{s}$ and $\mathbf{e}_{t}$ are input into a multi-modal fusion module (\eg MulT) for feature integration, yielding the final multi-modal prediction result:
\begin{equation}
y_{multi} = MultiModel \left ( \mathbf{e}_{s}, \mathbf{e}_{t};\mathbf{\theta} _{trainable}  \right ),
\end{equation}
where $MultiModel$ represents a classical multi-modal fusion method (\eg MulT), $\mathbf{\theta} _{trainable}$ denotes the trainable parameters, and $y_{multi}$ is the predicted output for the multi-modal classification task.

\begin{table*}[!t]
  \centering
  \caption{Details of The Selected Acoustic and Linguistic Foundation Models. Model size abbreviations: ``-B'': Base, ``-S'': Small, ``-M'': Medium, ``-L'': Large. ``-CN'': Pretraining using Chinese language corpora. Corpus abbreviations: ``LS'': LibriSpeech, ``LL'': LibriLight, ``MLS'': Multilingual Librispeech, ``VP'': VoxPopuli, ``CV'': CommonVoice, ``VL'': VoxLingua, ``BBL'': BABEL,``GS'': GigaSpeech,  ``MTD'': Multitask weakly supervised data, ``BC'': BookCorpus, ``OWT'': Open WebText, ``EW'': English Wikipedia,``CC'': Common Crawl, ``SE'': StackExchange, ``CX'': CulturaX, ``SNLI'': Stanford Natural Language Inference, ``MNLI'':  Multi-Genre Natural Language Inference.
}
  \label{tab:foundation_models}
  \begin{threeparttable}
      \begin{tabular}{lccrc}
        \toprule
         Model  &Pre-training Task  &Model Architecture &\# Params  &Training Datasets  \\
         \midrule
         \multicolumn{5}{c}{Acoustic} \\
         \midrule
        Wav2vec 2.0 B   &Contrastive Predictive Coding  &CNN + Transformer &95M  &LS-960  \\
        Wav2vec 2.0 L   &Contrastive Predictive Coding &CNN + Transformer &317M  &LS-960, LL-60k  \\
           &  & &  &VP-400K, MLS-50K,\\
        XLS-R 300M \tnote{*}   &Contrastive Predictive Coding &CNN + Transformer &300M  &CV, VL-107, BBL  \\
        HuBERT B   &Masked Prediction  &CNN + Transformer &95M  &LS-960  \\
        HuBERT L   &Masked Prediction  &CNN + Transformer &317M  &LS-960, LL-60k  \\
        Wav2vec 2.0 CN   &Masked Prediction  &CNN + Transformer &317M  &WenetSpeech \\
        HuBERT CN \tnote{\textdaggerdbl}  &Masked Prediction  &CNN + Transformer &317M  &WenetSpeech  \\
        ExHuBERT   & Fine-tuned based on HuBERT  &CNN + Transformer + MLP &380M  &EmoSet++  \\
        Data2vec B   &Masked Prediction, Self-distillation  &Transformer &93.4M  &LS-960  \\
        Data2vec L   &Masked Prediction, Self-distillation  &Transformer &314.2M  &LS-960, LL-60k  \\
        WavLM B   &Denoising Masked Speech Modeling  &CNN + Transformer &94.7M  &LS-960  \\
           &  & &  &LS-960, LL-60k,  \\
        WavLM L   &Denoising Masked Speech Modeling  &CNN + Transformer &316.6M  &GS-10K, VP-24K  \\
        Whisper S \tnote{*}   &Multi-task,ASR, Translation, etc.  &Encoder-Decoder Transformer &244M  &MTD-680k  \\
        Whisper M \tnote{*}  &Multi-task, ASR, Translation, etc.  &Encoder-Decoder Transformer &769M  &MTD-680k  \\
           &Contrastive learning,  & &  &MSP-Podcast, CMU-MOSEI \\
        Emotion2vec   & Masked Prediction, Self-distillation  &Transformer &94M  & IEMOCAP, MELD, MEAD  \\
        \midrule
         \multicolumn{5}{c}{Linguistic} \\
         \midrule
        RoBERTa L   &Masked Language Model  &Transformer &355M  &BC, EK, CCNews, OWT, Stories  \\ 
        DeBERTa L   &Masked Language Model  &Transformer &438M  &EW, BC, EK, OWT, Stories  \\ 
        XLNet L   &Permutation Language Model   &Transformer-XL &340M  &BC,EK, Giga5, ClueWeb, CC  \\ 
        SimCSE-   &   & &  &  \\ 
        RoBERTa L   &Contrastive Learning   &Transformer &355M  &EW, MNLI, SNLI  \\ 
           &   & &  &EW, Reddit, CC, SE \\
        E5 B   &Weakly-supervised Contrastive Learning   &Transformer &222M  &CCNews, S2ORC, Others  \\
           &   & &  &EW, Reddit, CC, SE  \\
        E5 L   &Weakly-supervised Contrastive Learning   &Transformer &605M  &CCNews, S2ORC, Others  \\
        GTE B   &Multi-stage Contrastive Learning   &Transformer &110M  &CC, Reddit, S2ORC, EW, SE, Others \\
        GTE L   &Multi-stage Contrastive Learning   &Transformer &330M  &CC, Reddit, S2ORC, EW, SE, Others \\
        mGTE B \tnote{*}  &Multilingual Contrastive Learning   &Transformer &264M  &C4, Skypile, mC4, CX, EW, Others  \\
        mE5 L \tnote{*}  &Weakly-supervised Contrastive Learning   &Transformer &922M  &EW, mC4, CCNews, S2ORC, Others \\
        Llama-3.1-8B   &Predict Next Token   &Transformer &8B  &15T Tokens  \\
        Baichuan2-7B \tnote{\textdaggerdbl}  &Predict Next Token   &Transformer &7B  &2.6T Tokens  \\
        Qwen2.5-7B \tnote{*}  &Predict Next Token   &Transformer &7B  &18T Tokens  \\
      \bottomrule
    \end{tabular}
    \begin{tablenotes}
        \item[*] Multilingual model.
        \item[\textdaggerdbl] Chinese-centric model. Unmarked models are English-centric.
    \end{tablenotes}
\end{threeparttable}
\vspace{-.3cm}
\end{table*}

\vspace{-.3cm}
\subsection{Acoustic Foundation Models}
This section introduces the acoustic foundational models used in this study. These models, pre-trained on large-scale speech datasets, extract rich speech representations and exhibit strong cross-task generalization. Table~\ref{tab:foundation_models} summarizes their key information.

\begin{itemize}
    \item \textbf{Wav2vec 2.0} \cite{baevski2020wav2vec}: Employs contrastive predictive coding and masked prediction to learn speech representations from raw audio.
    \item \textbf{XLS-R} \cite{babu2021xls}: Extends Wav2vec 2.0 with multi-lingual masked prediction, pre-trained on a large multi-lingual speech corpus.
    \item \textbf{HuBERT} \cite{hsu2021hubert}: Utilizes masked prediction of pseudo-labels generated via k-means over acoustic features.
    \item \textbf{WavLM} \cite{chen2022wavlm}: Introduces a masked speech denoising task to enhance performance on full-stack speech processing.
    \item \textbf{Data2vec} \cite{baevski2022data2vec}: Proposes a unified framework using masked prediction based on a teacher model's contextualized representations.
    \item \textbf{Emotion2vec} \cite{ma2023emotion2vec}: Designed for speech emotion representation by incorporating contrastive and prediction losses on emotional data.
    \item \textbf{ExHuBERT} \cite{amiriparian2024exhubert}: Enhances HuBERT with Backbone Block Expansion (BBE) for improved emotion recognition.
    \item \textbf{Whisper} \cite{radford2023robust}: A multi-task supervised model trained for large-scale speech recognition and translation.
\end{itemize}

\vspace{-.3cm}
\subsection{Linguistic Foundation Models}
This study explores advanced linguistic foundation models pre-trained on large-scale corpora to capture rich semantic information. Table~\ref{tab:foundation_models} summarizes their key information.

\begin{itemize}
    \item \textbf{RoBERTa} \cite{liu2019roberta}: A robustly optimized BERT approach with larger datasets and dynamic masking.
    \item \textbf{XLNet} \cite{yang2019xlnet}: A generalized autoregressive model that uses permutation language modeling to capture bidirectional context.
    \item \textbf{DeBERTa} \cite{he2020deberta}: Improves BERT with a disentangled attention mechanism and an enhanced mask decoder.
    \item \textbf{SimCSE} \cite{gao2021simcse}: A simple contrastive learning framework for learning high-quality sentence embeddings.
    \item \textbf{E5} \cite{wang2022text}: Employs weakly supervised contrastive pre-training on text pairs for effective text embeddings.
    \item \textbf{GTE} \cite{li2023towards}: A text embedding model trained with multi-stage contrastive learning.
    \item \textbf{mGTE} \cite{zhang2024mgte}: A multi-lingual version of GTE focused on text retrieval tasks.
    \item \textbf{mE5} \cite{wang2024multilingual}: A multi-lingual E5 model that provides sentence embeddings for numerous languages.
    \item \textbf{Baichuan 2} \cite{yang2023baichuan}: A large language model optimized for Chinese and English understanding and generation.
    \item \textbf{Llama 3} \cite{dubey2024llama}: A family of large language models known for their efficiency and strong performance.
    \item \textbf{Qwen2.5} \cite{yang2024qwen2}: A large language model emphasizing multi-lingual capabilities and practical application.
\end{itemize}

\vspace{-.3cm}
\subsection{Detection Tasks for Neuropsychiatric Disorders}
To instantiate and validate the FEND framework, we select three representative and highly prevalent neuropsychiatric disorders: AD, Depression, and ASD. The choice of these specific disorders is motivated by three key factors: (1) \textbf{Data Availability:} They have a relatively larger number of publicly available speech and text datasets, enabling large-scale, multi-corpus evaluation. (2) \textbf{Research Foundation:} There is a substantial body of existing research on their acoustic and linguistic markers, providing a solid basis for comparison. (3) \textbf{Representativeness:} They represent a diverse range of conditions, including a neurodegenerative disorder (AD), a mood disorder (Depression), and a neurodevelopmental disorder (ASD), making them ideal for testing the framework's versatility. While FEND is designed to be extensible to other neuropsychiatric conditions, these three serve as the initial focus of our investigation.

\begin{itemize}
    \item \textbf{AD}: AD manifests in speech and language as reduced fluency, increased grammatical errors, and limited vocabulary. The goal of AD detection is to identify linguistic abnormalities, providing insights into cognitive decline.
    \item \textbf{Depression}: Depression is characterized by abnormalities in speech and language features, such as slowed speech rate, monotonous intonation, and negative emotional content. The goal of depression detection is to identify these features using machine learning techniques, enabling timely intervention and personalized treatment.
    \item \textbf{ASD}: ASD is characterized by monotonous speech, abnormal prosody, and insufficient social interaction in language expression. The goal of ASD detection is to identify these communication and social abnormalities, aiding in early diagnosis and intervention.
\end{itemize}

\vspace{-.3cm}
\subsection{The Unified Evaluation Metrics}
Considering the diverse evaluation metrics used in prior approaches \cite{dong2025MHSDB, zhang2024paralbench, ma2024emobox}, we take unified evaluation metrics in FEND to facilitate fair and consistent comparisons across different neuropsychiatric disorder detection tasks. We encourage future researchers to adhere to these metrics, ensuring reproducibility and comparability of results.

We employ Weighted Accuracy (WA), Unweighted Accuracy (UA), and Weighted F1 Score (WF1) as evaluation metrics.  Here, WA corresponds to the standard overall accuracy. UA, also known as Balanced Accuracy, is the average of recall scores for each class, which treats each class equally and is thus suitable for imbalanced datasets. WF1 is the F1 score calculated for each class and then averaged, weighted by the number of true instances (support) for each class.

\begin{table*}[!t]
  \centering
  \caption{Details of the selected datasets over lifespan neuropsychiatric disorders (i.\,e., Alzheimer's disease, depression, and autism spectrum disorder).}
  \label{tab:datsets}
  \begin{threeparttable}
      \begin{tabular}{cccrcrrrrr}
        \toprule
        Lifespan  &Tasks  &Datasets   &\# Spk  &Language &Duration [h]  &\# Total &\# Training &\# Validation &\# Test\\
        \midrule
        Elderly    &Alzheimer  &PITT \cite{becker1994natural}      &292 &English  &10.68 &549  &384 &- &165\\
        Elderly &Alzheimer   &ADReSS \cite{luz2021alzheimer}    &156 &English  &3.26 &156  &108 &- &48\\
        Elderly    &Alzheimer  &ADReSSo \cite{luz2021detecting}     &237 &English  &5.07 &237  &166 &- &71\\
        Elderly    &Alzheimer  &ADReSS-M \cite{luz2023multilingual}    &281 &English \& Greek  &5.35 &281  &235 &- &46\\
        Elderly &Alzheimer &TAUKADIA \cite{luz2024connected}    &508 &English \& Chinese  &8.81 &508  &387 &- &121\\
           
        \midrule
        Adults   &Depression &DAIC-WOZ \cite{gratch2014distress}     &189 &English  &50.39 &189  &107 &35 &47\\
        Adults   &Depression  &E-DAIC \cite{devault2014simsensei}     &387 &English  &104.44 &387  &163 &56 &168\\
        Adults   &Depression  &D-VLOG \cite{yoon2022d}    &961 &English  &139.41 &961  &647 &102 &212\\
        Adults   &Depression  &EATD \cite{shen2022automatic}     &162 &Chinese  &2.15 &162  &83 &- &79\\
        Adults   &Depression &MODMA \cite{cai2022multi}     &52 &Chinese  &7.19 &52  &36 &- &16\\
        Adults   &Depression  &CMDC \cite{zou2022semi}    &78 &Chinese  &10.88 &78  &60 &- &18\\
        \midrule
        Children   &Autism  &CPSD \cite{schuller2013interspeech}    &99 &French  &1.04 &2,542  &903 &819 &820\\
        Children   &Autism  &ASDBank \cite{kuijper2015he}    &84 &Dutch  &3.96 &84  &58 &- &26\\
      \bottomrule
    \end{tabular}
  \end{threeparttable}
  \vspace{-.3cm}
\end{table*}

\section{EXPERIMENTAL SETUP}\label{sec:experimental}
This section describes the experimental setup, including the multi-modal datasets and implementation details. We detail the datasets for AD, depression, and ASD, covering their sources, characteristics, and partitioning. All datasets include speech modality, with Chinese audio transcribed using the SenseVoice \cite{an2024funaudiollm} and other languages using Whisper-Large \cite{radford2023robust}.

\vspace{-.3cm}
\subsection{Datasets}
For each dataset, we indicate whether the data splits are officially provided or manually partitioned by us. Table~\ref{tab:datsets} summarizes their key characteristics, including the number of speakers, language, duration, and data splits.

\subsubsection{AD} The AD datasets include PITT~\cite{becker1994natural}, ADReSS~\cite{luz2021alzheimer}, ADReSSo~\cite{luz2021detecting}, ADReSS-M~\cite{luz2023multilingual}, and TAUKADIAL~\cite{luz2024connected}. PITT contains English speech recordings from AD patients and healthy controls performing the ``Cookie Theft'' task, which we manually divided into training and test sets in a 7:3 ratio. ADReSS, ADReSSo, ADReSS-M, and TAUKADIAL are officially divided into training and test sets, with ADReSS-M presenting a demanding cross-lingual challenge, featuring training data primarily in English and test data in Greek, and TAUKADIAL including connected speech in English and Chinese.

\subsubsection{Depression} The depression datasets include DAIC-WOZ~\cite{gratch2014distress}, E-DAIC~\cite{devault2014simsensei}, D-VLOG~\cite{yoon2022d}, EATD~\cite{shen2022automatic}, MODMA~\cite{cai2022multi}, and CMDC~\cite{zou2022semi}. DAIC-WOZ and E-DAIC provide semi-structured English interviews, officially divided into training, validation, and test sets. D-VLOG includes English vlog videos, with access limited to 861 videos due to privacy issues. EATD contains Chinese interview audio and text, officially divided into training and test sets. MODMA and CMDC provide EEG and speech recordings, which we manually divided into training and test sets in a 7:3 ratio.

\subsubsection{ASD} The ASD datasets include CPSD~\cite{schuller2013interspeech} and ASDBank~\cite{kuijper2015he}. CPSD features French speech from children, officially divided into training, validation, and test sets. ASDBank contains Dutch spontaneous speech, which we manually divided into training and test sets in a 7:3 ratio.

\vspace{-.2cm}
\subsection{Implementation Details}
To ensure a fair comparison of different modality-specific foundation models, we employed a standardized training protocol across all experiments. For mono-modal experiments, a three-layer MLP served as the classification head. In multi-modal experiments, considering the typically extended duration of neuropsychiatric data samples (with the majority exceeding 60 seconds) and the requirement for sequence alignment in sequence-level fusion methods, we adopted a preprocessing and alignment strategy inspired by HAFFormer \cite{dong2024hafformer}. Specifically, representations from both speech and text modalities were mapped to 128 dimensions, and sequence lengths were downsampled to 200 via merging layer before fusion. All models were trained with an initial learning rate of 1e-3, a weight decay of 1e-5, for 80 epochs. A dropout rate of 0.2 was applied to the classification layer. This standardized training configuration enables a more objective assessment of various approaches, ensures comparability of experimental outcomes, and reveals inherent model differences, unaffected by variations in training conditions.

This uniform training configuration is a deliberate design choice, prioritizing comparability and reproducibility, which are central to a benchmark's integrity. To this end, our experimental paradigm is guided by two key principles: First, we employ a fixed set of hyperparameters across all experiments. Second, we adopt a feature-freezing approach, where the backbones of the foundation models remain untuned. The unified purpose of these principles is to scientifically isolate the performance of the pre-trained features---our primary object of study---from confounding variables such as hyperparameter optimization or task-specific fine-tuning adaptation. This ensures that the observed performance differences are attributable to the intrinsic representational power of the models themselves, thereby providing a fair and stable baseline for the community.

\section{RESULTS AND DISCUSSION}\label{sec:results}
This section provides a detailed analysis of the experimental results of the FEND framework for detecting AD,  depression, and ASD. We first present the results of mono-modal experiments, examining the performance of speech and text modalities across the three tasks, and then analyze the characteristics of different foundational models and datasets.

\vspace{-.3cm}
\subsection{Mono-Modal Benchmark}
We establish mono-modal benchmarks for all datasets in AD, depression, and ASD detection tasks, as summarized in Table \ref{tab:monomodal_AD}, Table \ref{tab:monomodal_depression} and Table \ref{tab:monomodal_asd}. These benchmarks provide guidance for feature selection and highlight the potential for developing robust feature extractors.

\subsubsection{AD Detection Results}
In the AD detection task, WavLM-Large achieves the best performance in the speech modality, with an average WF1 of 75.6\%. Its superior performance is possibly attributed to its large-scale pre-training and advanced denoising mechanisms \cite{chen2022wavlm}. As shown in Table \ref{tab:monomodal_AD}, WavLM-Large achieves WF1 scores of 80.6\% and 84.5\% on the PITT and ADReSS datasets, respectively.
In the text modality, E5-Base and E5-Large both perform well, but there are notable differences between them. E5-Large demonstrates stable performance across multiple monolingual (English) datasets, achieving WF1 of 82.3\%,  83.3\% and 87.3\% on the PITT, ADReSS and ADReSSo datasets, respectively. However, the performance of English-centric models like E5-Large significantly degrades on cross-lingual tasks. For instance, on the ADReSS-M dataset, where the training set is overwhelmingly English (with only 8 samples in Greek) and the test set is entirely Greek, its WF1 drops to 61.2\%. This performance degradation is expected, as the English-centric model has insufficient exposure to the target language. In contrast, the multilingual model mE5-L shows substantially better performance with a WF1 of 72.3\%, underscoring the importance of multilingual pre-training for cross-lingual generalization.

It is also noteworthy to compare the performance of generative Large Language Models (LLMs) like Llama-3.1-8B, Baichuan2-7B, and Qwen2.5-7B against specialized embedding models such as E5 and GTE. Our results indicate that these LLMs, utilized as feature extractors, exhibit strong and competitive performance across several datasets, likely benefiting from their extensive world knowledge and deep contextual understanding. However, they do not consistently outperform top-tier embedding-focused models like E5-Large, which are specifically optimized for generating high-quality semantic representations and often achieve comparable or superior results with greater efficiency. This suggests that while generative LLMs are a powerful asset for linguistic analysis, the choice between them and specialized embedding models involves a trade-off between broad knowledge and task-specific optimization, a direction that warrants further exploration within the FEND framework.

Overall, the text modality slightly outperforms the speech modality, with average WF1 scores of 73.4\% and 70.5\%, respectively, suggesting that text features may more effectively capture AD-related linguistic abnormalities.

\begin{table*}[!t]
  \centering
  \caption{Mono-Modal and multi-modal benchmark for AD detection. In this table, we report mono-modal and multi-modal results for all AD datasets under the same experimental setup. 
  Fusion methods: $\spadesuit$ denotes utterance-level, $\clubsuit$ denotes sequence-level.
  The \textunderscore\ denotes the top-performing mono-modal foundation model in overall performance. The $^{\uparrow}$ indicates multi-modal methods that surpass the best mono-modal performance.
  }
  \label{tab:monomodal_AD}
  \begin{threeparttable}
  \resizebox{\textwidth}{!}{
      \begin{tabular}{ccccccccccccccccccc}
        \toprule
         Model  &\multicolumn{3}{c}{PITT} &\multicolumn{3}{c}{ADReSS} &\multicolumn{3}{c}{ADReSSo} &\multicolumn{3}{c}{ADReSS-M}  &\multicolumn{3}{c}{TAUKADIA}  &\multicolumn{3}{c}{Mean}\\
               $[\%]$ &WA &UA	&WF1 &WA &UA &WF1 &WA &UA &WF1 &WA &UA &WF1 &WA &UA &WF1 &WA &UA &WF1  \\
         \midrule
         \multicolumn{19}{c}{Acoustic} \\
         \midrule
        Wav2vec 2.0 B &78.8 &78.1 	&78.7  &75.0 	&75.0 	&75.0 	&74.6 	&74.6 	&74.6 	&58.7 	&59.1 	&58.4 	&52.5 	&51.4 	&50.2 		&67.9 	&67.6 	&67.4 
    \\
        Wav2vec 2.0 L &78.2 	&78.0 	&78.2  &\textcolor{red!100}{87.5} 	&\textcolor{red!100}{87.5} 	&\textcolor{red!100}{87.5} 	&76.1 	&76.1 	&76.1 	&58.7 	&59.5 	&57.4 	&59.2 	&58.3 	&57.8 	 	&71.9 	&71.9 	&71.4 
\\
        XLS-R 300M  &66.1 	&63.9 	&64.7  &79.2 	&79.2 	&78.8 	&71.8 	&72.1 	&70.8 	&69.6 	&68.9 	&68.9 	&52.5 	&52.4 	&52.5 		&67.8 	&67.3 	&67.1 
\\
        HuBERT B  &78.8 	&78.6 	&78.8  &72.9 	&72.9 	&72.8 	&80.3 	&80.3 	&80.3 	&58.7 	&59.5 	&57.4 	&\textcolor{red!100}{62.5} 	&\textcolor{red!70}{61.8} 	&\textcolor{red!100}{61.7} 	 	&70.6 	&70.6 	&70.2 
 \\
        HuBERT L &\textcolor{red!100}{83.6} 	&\textcolor{red!70}{82.8} 	&\textcolor{red!100}{83.5} &79.2 	&79.2 	&79.1 	&\textcolor{red!45}{81.7} 	&\textcolor{red!45}{81.5} 	&\textcolor{red!45}{81.5} 	&67.4 	&67.4 	&67.4 	&56.7 	&56.5 	&56.6 	 	&73.7 	&73.5 	&73.6 
  \\
        ExHuBERT    &73.3 	&72.3 	&73.0  &68.8 	&68.8 	&68.4 	&67.6 	&67.6 	&67.6 	&65.2 	&65.5 	&65.1 	&55.8 	&55.8 	&55.8 	 	&66.1 	&66.0 	&66.0 
  \\
        Data2vec B  &80.0 	&80.4 	&80.1 &83.3 	&83.3 	&83.3 	&78.9 	&78.8 	&78.9 	&65.2 	&65.2 	&65.2 	&51.7 	&50.8 	&50.2 	 	&71.8 	&71.7 	&71.5 
 \\
        Data2vec L  &73.9 	&73.9 	&74.0  &\textcolor{red!70}{85.4} 	&\textcolor{red!70}{85.4} 	&\textcolor{red!45}{85.3} 	&76.1 	&75.9 	&75.8 	&65.2 	&63.8 	&61.0 	&54.2 	&53.7 	&53.7 &71.0 	&70.5 	&70.0 
\\
        WavLM B &75.8 	&75.7 	&75.8   &70.8 	&70.8 	&70.8 	&78.9 	&78.7 	&78.5 	&69.6 	&69.3 	&69.3 	&54.2 	&53.8 	&53.9 	&69.9 	&69.7 	&69.7 
\\
        \underline{WavLM L} &\textcolor{red!45}{80.6} 	&\textcolor{red!45}{80.5} 	&\textcolor{red!45}{80.6}  &81.2 	&81.2 	&81.2 	&\textcolor{red!100}{84.5} 	&\textcolor{red!100}{84.4} 	&\textcolor{red!100}{84.5} 	&\textcolor{red!70}{71.7} 	&71.2	&71.3 	&60.8 	&60.4 	&60.5 	 	&\textcolor{red!100}{75.8} 	&\textcolor{red!100}{75.5} 	&\textcolor{red!100}{75.6} 
 \\
        Whisper S &80.0 	&\textcolor{red!45}{80.5} 	&80.1  &77.1 	&77.1 	&76.8 	&76.1 	&75.9 	&75.8 	&\textcolor{red!70}{71.7} 	&\textcolor{red!70}{71.8} 	&\textcolor{red!70}{71.8} 	&\textcolor{red!70}{61.7} 	&\textcolor{red!45}{61.0} 	&\textcolor{red!70}{60.9} 	&73.3 	&73.3 	&73.1  
   \\
        Whisper M  &75.2 	&76.2 	&75.2 &79.2 	&79.2 	&79.2 	&71.8 	&71.7 	&71.6 	&52.2 	&52.8 	&51.1 	&55.8 	&55.6 	&55.8 	 	&66.8 	&67.1 	&66.6 
  \\
        Emotion2vec &\textcolor{red!70}{82.4} 	&\textcolor{red!100}{83.0} 	&\textcolor{red!70}{82.5}  &\textcolor{red!70}{85.4} 	&\textcolor{red!70}{85.4} 	&\textcolor{red!70}{85.4} 	&\textcolor{red!70}{83.1} 	&\textcolor{red!70}{83.0} 	&\textcolor{red!70}{83.0} 	&\textcolor{red!70}{71.7} 	&\textcolor{red!70}{71.8} 	&\textcolor{red!70}{71.8} 	&52.5 	&51.8 	&51.7 	 	&\textcolor{red!70}{75.0} 	&\textcolor{red!70}{75.0} 	&\textcolor{red!70}{74.9} 
  \\
        Wav2vec 2.0 CN  &74.5 	&73.6 	&74.3   &68.8 	&68.8 	&68.1 	&73.2 	&73.0 	&72.2 	&67.4 	&67.4 	&67.4 	&50.0 	&49.0 	&48.1 	 	&66.8 	&66.4 	&66.0 
 \\
        HuBERT CN   	&79.4 	&78.3 	&79.1   &81.2 	&81.2 	&81.2 	&76.1 	&76.1 	&76.1 	&\textcolor{red!100}{73.9} 	&\textcolor{red!100}{74.2} 	&\textcolor{red!100}{73.8} 	&\textcolor{red!100}{62.5} 	&\textcolor{red!100}{63.6} 	&\textcolor{red!45}{60.7}  	&\textcolor{red!45}{74.6} 	&\textcolor{red!45}{74.7} 	&\textcolor{red!45}{74.2} 
   \\
        \midrule
         \multicolumn{19}{c}{Linguistic} \\
         \midrule
        RoBERTa L  &80.0 	&80.5 	&80.1  &79.2 	&79.2 	&79.2 	&\textcolor{red!100}{88.7} 	&\textcolor{red!100}{88.7} 	&\textcolor{red!100}{88.7} 	&60.9  &61.7 	&59.3 	&58.3 	&57.6 	&57.5 	&73.4 	&73.5 	&73.0  
 \\ 
        DeBERTa L &\textcolor{red!70}{81.8} 	&\textcolor{red!100}{82.4} 	&\textcolor{red!45}{81.9}  &81.2 	&81.2 	&81.2 	&84.5 	&84.5 	&84.5 	&67.4 	&66.9 	&66.8 	&60.0 	&59.8 	&60.0 	 	&75.0 	&75.0 	&\textcolor{red!45}{74.9}  
 \\ 
        XLNet L  &78.8 	&78.1 	&78.7  &81.2 	&81.2 	&81.2 	&83.1 	&83.1 	&83.1 	&56.5 	&56.6 	&56.5 	&60.0 	&61.3 	&57.2 		&71.9 	&72.1 	&71.3 
   \\ 
        SimCSE-     \\
        RoBERTa L 	&78.2 	&78.6 	&78.3   &77.1 	&77.1 	&77.0 	&\textcolor{red!45}{85.9} 	&85.8 	&85.7 	&54.3 	&54.5 	&54.3 	&\textcolor{red!70}{63.3} 	&\textcolor{red!70}{63.2} 	&\textcolor{red!70}{63.3} 	&71.8 	&71.8 	&71.7 
\\ 
        E5 B  	&77.6 	&77.1 	&77.5  &\textcolor{red!70}{83.3} 	&\textcolor{red!70}{83.3} 	&\textcolor{red!70}{83.3} 	&\textcolor{red!45}{85.9} 	&\textcolor{red!45}{86.0} 	&\textcolor{red!45}{85.9} 	&\textcolor{red!70}{69.6} 	&\textcolor{red!70}{69.3} 	&\textcolor{red!70}{69.4} 	&\textcolor{red!100}{64.2} 	&\textcolor{red!100}{63.5} 	&\textcolor{red!100}{63.4}  	&\textcolor{red!100}{76.1} 	&\textcolor{red!100}{75.8} 	&\textcolor{red!100}{75.9} 
\\
        \underline{E5 L} &\textcolor{red!100}{82.4} 	&\textcolor{red!70}{81.8} 	&\textcolor{red!70}{82.3} &\textcolor{red!70}{83.3} 	&\textcolor{red!70}{83.3} 	&\textcolor{red!70}{83.3} 	&\textcolor{red!70}{87.3} 	&\textcolor{red!70}{87.3} 	&\textcolor{red!70}{87.3} 	&63.0 	&62.1 	&61.2 	&60.0 	&59.8 	&60.0 	 	&\textcolor{red!45}{75.2} 	&74.9 	&74.8 
  \\
        GTE B   &78.8 	&80.4 	&78.6 &79.2 	&79.2 	&79.1 	&83.1 	&83.0 	&83.0 	&56.5 	&56.3 	&56.4 	&61.7 	&60.4 	&58.9 	 	&71.9 	&71.9 	&71.2
 \\
        GTE L   &78.8 	&79.1 	&78.9  &\textcolor{red!100}{85.4} 	&\textcolor{red!100}{85.4} 	&\textcolor{red!100}{85.3} 	&\textcolor{red!45}{85.9} 	&85.9 	&\textcolor{red!45}{85.9} 	&67.4 	&67.4 	&67.4 	&60.0 	&60.1 	&60.0 		&\textcolor{red!70}{75.5} 	&\textcolor{red!70}{75.6} 	&\textcolor{red!70}{75.5} 
   \\
        mGTE B  &78.2 	&78.5 	&78.3  &79.2 	&79.2 	&79.2 	&83.1 	&83.1 	&83.1 	&63.0 	&63.6 	&62.4 	&56.7 	&56.5 	&56.6 	 	&72.0 	&72.2 	&71.9 
 \\
        mE5 L 	&79.4 	&79.7 	&79.5  &79.2 	&79.2 	&79.2 	&84.5 	&84.5 	&84.5 	&\textcolor{red!100}{73.9} 	&\textcolor{red!100}{72.9} 	&\textcolor{red!100}{72.3} 	&58.3 	&59.0 	&57.7  	&75.1 	&\textcolor{red!45}{75.1} 	&74.6 

\\
        Llama-3.1-8B  &80.6 	&80.8 	&80.7   &79.2 	&79.2 	&79.2 	&\textcolor{red!45}{85.9} 	&85.9 	&\textcolor{red!45}{85.9} 	&63.0 	&63.1 	&63.1 	&56.7 	&56.6 	&56.7 	 	&73.1 	&73.1 	&73.1 
 \\
        Baichuan2-7B &\textcolor{red!100}{82.4} 	&\textcolor{red!100}{82.4} 	&\textcolor{red!100}{82.5}  &79.2 	&79.2 	&79.2 	&\textcolor{red!45}{85.9} 	&86.0 	&\textcolor{red!45}{85.9} 	&63.0 	&62.1 	&61.2 	&58.3 	&57.6 	&57.3 		&73.8 	&73.5 	&73.2 
 \\
        Qwen2.5-7B &77.0 	&77.1 	&77.0  &77.1 	&77.1 	&76.8 	&83.1 	&83.1 	&83.1 	&\textcolor{red!70}{69.6} 	&\textcolor{red!45}{68.8} 	&\textcolor{red!45}{68.4} 	&\textcolor{red!70}{63.3} 	&\textcolor{red!45}{62.7} 	&\textcolor{red!45}{62.6} 		&74.0 	&73.8 	&73.6 
 \\


\midrule
         \multicolumn{19}{c}{Multi-Model Fusion} \\
         \midrule
        $\clubsuit$MFM  	&80.0 	&80.4 	&80.1  &50.0 	&50.0 	&33.3 	&81.7 	&81.5 	&81.5 	&\textcolor{red!45}{63.0} 	&\textcolor{red!45}{62.1} 	&\textcolor{red!45}{61.2} 	&55.8 	&54.9 	&54.3  	&66.1 	&65.8 	&62.1 
\\
        $\clubsuit$MFN  &\textcolor{red!70}{81.8} 	&\textcolor{red!45}{82.3} 	&\textcolor{red!70}{81.9}  &62.5 	&62.5 	&62.5 	&\textcolor{red!100}{87.3}$^{\uparrow}$ 	&\textcolor{red!100}{87.3}$^{\uparrow}$ 	&\textcolor{red!100}{87.3}$^{\uparrow}$ 	&56.5 	&56.1 	&56.0 	&\textcolor{red!100}{63.3}$^{\uparrow}$ 	&\textcolor{red!100}{63.6}$^{\uparrow}$ 	&\textcolor{red!100}{63.3}$^{\uparrow}$ 		&70.3 	&70.4 	&70.2 
 \\
        $\clubsuit$GraphMFN  &64.2 	&64.1 	&64.3 &\textcolor{red!100}{85.4}$^{\uparrow}$ 	&\textcolor{red!100}{85.4}$^{\uparrow}$ 	&\textcolor{red!100}{85.4}$^{\uparrow}$ 	&78.9 &	79.0 	&78.8 	&56.5 	&54.9 	&49.5 	&\textcolor{red!70}{60.0} 	&\textcolor{red!70}{59.9} 	&\textcolor{red!70}{60.0} 	 	&69.0 	&68.7 	&67.6 
 \\
        $\clubsuit$MulT  &76.4 	&75.7 	&76.3 &62.5 	&62.5 	&62.4 	&78.9 	&78.9 	&78.9 	&52.2 	&54.0  &42.6 	&55.8 	&56.3 	&55.4 	 	&65.2 	&65.5 	&63.1 
 \\
        $\spadesuit$MMIM  &\textcolor{red!70}{81.8} 	&\textcolor{red!70}{82.7} 	&\textcolor{red!70}{81.9}  &81.2 	&81.2 	&81.2 	&84.5 	&84.4 	&84.4 	&\textcolor{red!100}{67.4} 	&\textcolor{red!100}{67.2} &\textcolor{red!100}{67.3} 	&\textcolor{red!45}{59.2} 	&\textcolor{red!45}{59.2} 	&\textcolor{red!45}{59.2} 	 	&\textcolor{red!100}{74.8} 	&\textcolor{red!100}{74.9} 	&\textcolor{red!100}{74.8} 
 \\
        $\spadesuit$LMF  &\textcolor{red!100}{84.2}$^{\uparrow}$ 	&\textcolor{red!100}{84.3}$^{\uparrow}$ 	&\textcolor{red!100}{84.3}$^{\uparrow}$   &\textcolor{red!100}{85.4}$^{\uparrow}$ 	&\textcolor{red!100}{85.4}$^{\uparrow}$ 	&\textcolor{red!100}{85.4}$^{\uparrow}$ 	&\textcolor{red!70}{85.9} 	&\textcolor{red!45}{85.8} 	&\textcolor{red!70}{85.9} 	&60.9 	&59.3 	&54.6 	&53.3 	&52.5 	&51.9 		&\textcolor{red!70}{73.9} 	&\textcolor{red!45}{73.5} 	&\textcolor{red!45}{72.4} 
 \\
        $\spadesuit$TFN   &81.2 	&80.7 	&81.2  &\textcolor{red!100}{85.4}$^{\uparrow}$ 	&\textcolor{red!100}{85.4}$^{\uparrow}$ 	&\textcolor{red!100}{85.4}$^{\uparrow}$ 	&\textcolor{red!70}{85.9} 	&\textcolor{red!70}{86.0} 	&\textcolor{red!70}{85.9} 	&\textcolor{red!45}{63.0} 	&\textcolor{red!45}{62.1} 	&\textcolor{red!45}{61.2} 	&58.3 	&56.8 	&53.9 		&\textcolor{red!100}{74.8} 	&\textcolor{red!70}{74.2} 	&\textcolor{red!70}{73.5} 
 \\
       $\spadesuit$Attention  &80.6 	&80.5 	&80.6    &77.1 	&77.1 	&77.1 	&81.7 	&81.6 	&81.6 	&\textcolor{red!70}{65.2} 	&\textcolor{red!70}{64.6} 	&\textcolor{red!70}{64.4} 	&56.7 	&56.1 	&56.2 		&72.3 	&72.0 	&72.0 
 \\
 
      \bottomrule
    \end{tabular}
    }
\end{threeparttable}
\end{table*}

\subsubsection{Depression Detection Results}
In the depression detection task, the speech and text modalities perform similarly, with the speech modality having a slight edge. WavLM-Large again leads in the speech modality, achieving an average WF1 of 81.2\%. As shown in Table \ref{tab:monomodal_depression}, it achieves WF1 scores of 87.8\% and 100\% on the D-VLOG and CMDC datasets, respectively.
In the text modality, E5-Large and mGTE-Base stand out. E5-Large, an English-centric model, achieves WF1 scores of 92.6\% and 88.2\% on the English D-VLOG and Chinese EATD datasets, respectively, benefiting from its weakly supervised contrastive learning pre-training strategy \cite{wang2022text}. Meanwhile, the multilingual mGTE-Base shows strong cross-lingual generalization, achieving a high WF1 of 88.9\% on the Chinese CMDC dataset, outperforming many English-centric models on non-English data.

Overall, the text modality slightly outperforms the speech modality in depression detection, with average WF1 scores of 76.7\% and 76.6\%, respectively. Although the difference is marginal, this result suggests that text features may more effectively capture subtle depression-related cues in this task.

\begin{table*}[!t]
  \centering
  \caption{Mono-modal and multi-modal benchmark for Depression detection. In this table, we report mono-modal and multi-modal results for all Depression datasets under the same experimental setup. 
  The definitions of all markers are the same as in Table~\ref{tab:monomodal_AD}.}
  \label{tab:monomodal_depression}
  \begin{threeparttable}
  \resizebox{\textwidth}{!}{
      \begin{tabular}{cccccccccccccccccccccc}
        \toprule
         Model  &\multicolumn{3}{c}{DAIC-WOZ} &\multicolumn{3}{c}{Extend DAIC} &\multicolumn{3}{c}{D-VLOG}  &\multicolumn{3}{c}{EATD} &\multicolumn{3}{c}{MODMA} &\multicolumn{3}{c}{CMDC} &\multicolumn{3}{c}{Mean}\\
                $[\%]$ &WA &UA &WF1 &WA &UA &WF1 &WA &UA &WF1 &WA &UA &WF1 &WA &UA &WF1 &WA &UA &WF1 &WA &UA &WF1 \\
         \midrule
         \multicolumn{22}{c}{Acoustic} \\
         \midrule
        Wav2vec 2.0 B &62.9 	&49.8 	&54.8 	&78.6 	&50.0 	&69.1 	&77.8 	&77.4 	&77.7 	&\textcolor{red!70}{86.1} 	&50.0 	&79.6 	&76.0 	&74.5 	&75.5 	&\textcolor{red!100}{100.0} 	&\textcolor{red!100}{100.0} 	&\textcolor{red!100}{100.0} 	&80.2 	&67.0 	&76.1 
    \\
        Wav2vec 2.0 L &65.7 	&52.0 	&56.5 	&78.6 	&50.0 	&69.1 	&82.0 	&81.1 	&81.7 	&\textcolor{red!100}{87.3} 	&\textcolor{red!45}{54.5} 	&\textcolor{red!45}{82.5} 	&78.1 	&76.9 	&77.8 	&88.9 	&91.7 	&89.2 	&80.1 	&67.7	&76.1 
\\
        XLS-R 300M  &65.7 	&58.0 	&63.7 	&73.2 	&\textcolor{red!100}{64.8} 	&\textcolor{red!45}{74.2}	&72.0 	&70.6 	&71.2 	&\textcolor{red!70}{86.1} 	&50.0 	&79.6 	&\textcolor{red!100}{83.3} 	&\textcolor{red!100}{83.6} 	&\textcolor{red!100}{83.4} 	&\textcolor{red!100}{100.0} 	&\textcolor{red!100}{100.0} 	&\textcolor{red!100}{100.0} &80.1 	&\textcolor{red!45}{71.2}	&78.7 
 \\
        HuBERT B  &68.6 	&54.2 	&58.3 	&\textcolor{red!100}{80.4} 	&54.2 	&73.1 	&79.4 	&78.4 	&79.0 	&\textcolor{red!70}{86.1} 	&50.0 	&79.6 	&\textcolor{red!70}{82.3} 	&\textcolor{red!45}{80.8} 	&\textcolor{red!45}{81.9} 	&\textcolor{red!100}{100.0} 	&\textcolor{red!100}{100.0} 	&\textcolor{red!100}{100.0} 	&\textcolor{red!45}{82.8} 	&69.6 	&78.7 
 \\
        HuBERT L  &68.6 	&\textcolor{red!100}{64.1} 	&\textcolor{red!70}{68.2} 	&\textcolor{red!100}{80.4} 	&\textcolor{red!45}{60.2} 	&\textcolor{red!100}{77.0} 	&\textcolor{red!45}{84.7} 	&\textcolor{red!45}{84.8} 	&\textcolor{red!45}{84.7} 	&\textcolor{red!70}{86.1} 	&50.0 	&79.6 	&72.9 	&73.3 	&73.0 	&88.9 	&87.5 	&88.9 	&80.3 	&70.0 	&78.6 
  \\
        ExHuBERT &57.1 	&49.5 	&55.4 	&67.9 	&\textcolor{red!70}{64.4} 	&70.3 	&79.9 	&79.3 	&79.8 	&\textcolor{red!70}{86.1} 	&53.8 	&81.7 	&71.9 	&71.8 	&72.0 	&83.3 	&83.3 	&83.6 	&74.4 	&67.0 	&73.8 
  \\
        Data2vec B  &65.7 	&50.0 	&52.1 	&78.6 	&50.0 	&69.1 	&81.0 	&80.6 	&80.9 	&\textcolor{red!70}{86.1} 	&50.0 	&79.6 	&66.7 	&67.5 	&66.7 	&83.3 	&83.3 	&83.6 	&76.9 	&63.6 	&72.0 
 \\
        Data2vec L &65.7 	&50.0 	&52.1 	&78.6 	&50.0 	&69.1 	&75.1 	&74.2 	&74.8 	&\textcolor{red!70}{86.1} 	&50.0 	&79.6 	&76.0 	&74.7 	&75.7 	&83.3 	&87.5 	&83.8 	&77.5 	&64.4 	&72.5 
 \\
        WavLM B &\textcolor{red!100}{74.3} 	&\textcolor{red!70}{62.5} 	&\textcolor{red!100}{68.7} 	&\textcolor{red!100}{80.4} 	&54.2	&73.1 	&82.0 	&81.9 	&82.0 	&\textcolor{red!70}{86.1} 	&50.0 	&79.6 	&\textcolor{red!70}{82.3} 	&\textcolor{red!70}{82.1} 	&\textcolor{red!70}{82.3} 	&88.9 	&91.7 	&89.2 	&82.3 	&70.4 	&79.2 
  \\
        \underline{WavLM L} &\textcolor{red!70}{71.4} 	&58.3 	&63.8 	&78.6 	&59.1 	&74.1 	&\textcolor{red!100}{87.8} 	&\textcolor{red!100}{87.8} 	&\textcolor{red!100}{87.8} 	&\textcolor{red!100}{87.3} 	&\textcolor{red!45}{54.5} 	&\textcolor{red!45}{82.5} 	&\textcolor{red!45}{79.2} 	&79.1 	&79.2 	&\textcolor{red!100}{100.0} 	&\textcolor{red!100}{100.0} 	&\textcolor{red!100}{100.0} 	&\textcolor{red!100}{84.1} 	&\textcolor{red!100}{73.1} 	&\textcolor{red!100}{81.2} 
 \\
        Whisper S &\textcolor{red!70}{71.4} 	&58.3 	&63.8 	&78.6 	&50.0 	&69.1 	&83.1 	&83.2 	&83.1 	&\textcolor{red!70}{86.1} 	&\textcolor{red!100}{69.1} 	&\textcolor{red!100}{85.8} 	&74.0 	&73.4 	&73.9 	&77.8 	&79.2 	&78.4 	&78.5 	&68.9 	&75.7  
   \\
        Whisper M &65.7 	&54.0 	&59.7 	&\textcolor{red!100}{80.4} 	&57.2 	&\textcolor{red!70}{75.4} 	&76.7 	&77.0 	&76.8 	&\textcolor{red!70}{86.1} 	&50.0 	&79.6 	&66.7 	&66.7 	&66.8 	&77.8 	&79.2 	&78.4 	&75.6 	&64.0 	&72.8 
  \\
        Emotion2vec &\textcolor{red!70}{71.4} 	&58.3 	&63.8 	&\textcolor{red!100}{80.4} 	&54.2 	&73.1 	&\textcolor{red!70}{85.7} 	&\textcolor{red!70}{85.4} 	&\textcolor{red!70}{85.7} 	&\textcolor{red!70}{86.1} 	&50.0 	&79.6 	&76.0 	&75.8 	&76.1 	&\textcolor{red!100}{100.0} 	&\textcolor{red!100}{100.0} 	&\textcolor{red!100}{100.0} 	&\textcolor{red!70}{83.3} 	&\textcolor{red!45}{70.6} 	&\textcolor{red!70}{79.7} 
  \\
        Wav2vec 2.0 CN &60.0 	&49.6 	&55.6 	&78.6 	&50.0 	&69.1 	&69.8 	&68.1 	&68.5 	&\textcolor{red!70}{86.1} 	&50.0 	&79.6 	&75.0 	&75.4 	&75.1 	&\textcolor{red!100}{100.0} 	&\textcolor{red!100}{100.0} 	&\textcolor{red!100}{100.0} 	&78.3 	&65.5 	&74.7 
 \\
        HuBERT CN  &\textcolor{red!70}{71.4} 	&\textcolor{red!45}{60.3} 	&\textcolor{red!45}{66.4} 	&78.6 	&56.1 	&74.1 	&74.6 	&75.0 	&74.6 	&84.8 	&\textcolor{red!70}{68.3} 	&\textcolor{red!70}{84.8} 	&78.1 	&78.2 	&78.2 	&\textcolor{red!100}{100.0} 	&\textcolor{red!100}{100.0} 	&\textcolor{red!100}{100.0}  &81.3 	&\textcolor{red!70}{73.0} 	&\textcolor{red!70}{79.7} 
   \\
        \midrule
         \multicolumn{22}{c}{Linguistic} \\
         \midrule
        RoBERTa L &\textcolor{red!70}{71.4} 	&60.3 	&66.4 	&76.8 	&51.9 	&70.9 	&89.9 	&90.4 	&90.0 	&86.1 	&50.0 	&79.6 	&63.5 	&62.6 	&63.4 	&\textcolor{red!70}{83.3} 	&\textcolor{red!70}{83.3} 	&\textcolor{red!45}{83.6} 	&78.5 	&66.4 	&75.7 
 \\ 
        DeBERTa L &\textcolor{red!70}{71.4} 	&62.3 	&\textcolor{red!45}{68.3} 	&76.8 	&54.9 	&72.9 	&88.4 	&88.6 	&88.4 	&\textcolor{red!70}{87.3} 	&\textcolor{red!45}{62.2} 	&\textcolor{red!70}{85.2} 	&75.0 	&73.5 	&74.5 	&77.8 	&79.2 	&78.4 	&79.5 	&70.1 	&78.0 
 \\ 
        XLNet L  &68.6 	&\textcolor{red!45}{64.1} 	&68.2 	&\textcolor{red!70}{80.4} 	&\textcolor{red!100}{63.3} 	&\textcolor{red!100}{78.3} 	&89.4 	&89.8 	&89.4 	&86.1 	&50.0 	&79.6 	&62.5 	&60.8 	&61.8 	&72.2 	&75.0 	&73.0 	&76.5 	&67.2 	&75.1 

   \\ 
        SimCSE-     \\
        RoBERTa L &68.6 	&54.2 	&58.3 	&71.4 	&\textcolor{red!70}{60.6} 	&72.2 	&88.9 	&89.0 	&88.9 	&86.1 	&50.0 	&79.6 	&67.7 	&66.3 	&67.2 	&77.8 	&79.2 	&78.4 	&76.8 	&66.6 	&74.1 
 \\ 
        E5 B &68.6 	&56.2 	&61.7 	&75.0 	&50.8 	&69.8 	&\textcolor{red!70}{92.1} 	&\textcolor{red!70}{92.1} 	&\textcolor{red!70}{92.1} 	&86.1 	&61.4 	&\textcolor{red!45}{84.3} 	&67.7 	&66.8 	&67.5 	&77.8 	&79.2 	&78.4 	&77.9 	&67.8 	&75.6 
\\
        \underline{E5 L} &\textcolor{red!100}{74.3} 	&\textcolor{red!100}{64.5} 	&\textcolor{red!100}{70.7} 	&78.6 	&59.1 	&75.7 	&\textcolor{red!100}{92.6} 	&\textcolor{red!100}{92.7} 	&\textcolor{red!100}{92.6} 	&\textcolor{red!100}{89.9} 	&\textcolor{red!100}{67.4} 	&\textcolor{red!100}{88.2} 	&65.6 	&63.6 	&64.6 	&77.8 	&79.2 	&78.4 	&79.8 	&\textcolor{red!45}{71.1} 	&\textcolor{red!45}{78.4} 
  \\
        GTE B  &62.9 	&55.8 	&61.3 	&\textcolor{red!70}{80.4} 	&54.2 	&73.1 	&87.3 	&88.1 	&87.3 	&83.5 	&\textcolor{red!70}{63.8} 	&83.2 	&64.6 	&61.6 	&62.3 	&72.2 	&75.0 	&73.0 	&75.2 	&66.4 	&73.4 
 \\
        GTE L  &65.7 	&58.0 	&63.7 	&\textcolor{red!70}{80.4} 	&\textcolor{red!45}{60.2} 	&\textcolor{red!70}{77.0} 	&\textcolor{red!45}{91.5} 	&91.7 	&91.5 	&81.0 	&58.5 	&80.6 	&59.4 	&58.3 	&59.2 	&72.2 	&75.0 	&73.0 	&75.0 	&67.0 	&74.2 
   \\
        mGTE B  &68.6 	&60.1 	&66.0 	&\textcolor{red!70}{80.4} 	&57.2 	&75.4 	&\textcolor{red!45}{91.5} 	&91.8 	&\textcolor{red!45}{91.6} 	&86.1 	&53.8 	&81.7 	&\textcolor{red!70}{78.1} 	&\textcolor{red!70}{78.2} 	&\textcolor{red!70}{78.2} 	&\textcolor{red!100}{88.9} 	&\textcolor{red!100}{87.5} 	&\textcolor{red!100}{88.9} 	&\textcolor{red!100}{82.3} 	&\textcolor{red!70}{71.4} 	&\textcolor{red!100}{80.3} 
 \\
        mE5 L  &60.0 	&57.6 	&60.6 	&\textcolor{red!70}{80.4} 	&54.2 	&73.1 	&89.4 	&89.5 	&89.4 	&84.8 	&53.1 	&80.9 	&69.8 	&70.2 	&69.9 	&\textcolor{red!70}{83.3} 	&\textcolor{red!100}{87.5} 	&\textcolor{red!70}{83.8} 	&78.0 	&68.7 	&76.3 
  \\
        Llama-3.1-8B &68.6 	&58.2 	&64.2 	&78.6 	&50.0 	&69.1 	&90.5 	&90.8 	&90.5 	&\textcolor{red!70}{87.3} 	&54.5 	&82.5 	&74.0 	&73.4 	&73.9 	&\textcolor{red!70}{83.3} 	&\textcolor{red!70}{83.3} 	&\textcolor{red!45}{83.6} 	&80.4 	&68.4 	&77.3 
  \\
        Baichuan2-7B &\textcolor{red!70}{71.4} 	&\textcolor{red!70}{64.3} 	&\textcolor{red!70}{69.7} 	&78.6 	&50.0 	&69.1 	&\textcolor{red!45}{91.5} 	&\textcolor{red!45}{91.9} 	&\textcolor{red!45}{91.6} 	&86.1 	&53.8 	&81.7 	&\textcolor{red!100}{80.2} 	&\textcolor{red!100}{79.5} 	&\textcolor{red!100}{80.1} 	&77.8 	&79.2 	&78.4 	&\textcolor{red!45}{80.9} 	&69.8 	&\textcolor{red!45}{78.4} 
  \\
        Qwen2.5-7B  &68.6 	&62.1 	&67.3 	&\textcolor{red!100}{82.1} 	&58.3 	&\textcolor{red!45}{76.7} 	&90.5 	&90.9 	&90.5 	&\textcolor{red!70}{87.3} 	&58.4 	&84.1 	&\textcolor{red!70}{78.1} 	&\textcolor{red!70}{78.2} 	&\textcolor{red!70}{78.2} 	&\textcolor{red!70}{83.3} 	&\textcolor{red!70}{83.3} 	&\textcolor{red!45}{83.6} 	&\textcolor{red!70}{81.7} 	&\textcolor{red!100}{71.9} 	&\textcolor{red!70}{80.1}  
 \\

          \midrule
         \multicolumn{22}{c}{Multi-Model Fusion} \\
         \midrule
        $\clubsuit$MFM  &65.7 	&50.0 	&52.1 	&\textcolor{red!100}{80.4} 	&54.2 	&73.1 	&\textcolor{red!70}{92.1} 	&\textcolor{red!70}{92.4} 	&\textcolor{red!70}{92.1} 	&86.1	&\textcolor{red!100}{72.9}	&\textcolor{red!100}{86.3}	&56.2	&50.0	&40.5	&83.3	&83.3	&83.6	&77.3 	&67.1 	&71.3 
\\
        $\clubsuit$MFN  &62.9 	&55.8 	&61.3 	&\textcolor{red!70}{78.6} 	&53.0 	&72.0 	&89.4 	&89.3 	&89.4 	&86.1	&50	&79.6	&68.8	&70.4	&68.5	&\textcolor{red!70}{94.4}	&\textcolor{red!70}{95.8}	&\textcolor{red!70}{94.5}	&80.0 	&69.1 	&77.6 
 \\
        $\clubsuit$GraphMFN  &\textcolor{red!100}{74.3}$^{\uparrow}$ 	&\textcolor{red!100}{72.5}$^{\uparrow}$ 	&\textcolor{red!100}{74.5}$^{\uparrow}$ 	&\textcolor{red!70}{78.6}$^{\uparrow}$ 	&\textcolor{red!100}{68.2}$^{\uparrow}$ 	&\textcolor{red!100}{78.6}$^{\uparrow}$ 	&90.5 	&90.2 	&90.5 	&\textcolor{red!100}{87.3}	&\textcolor{red!70}{62.2}	&\textcolor{red!45}{85.2}	&63.5	&62.0	&63.0	&77.8	&83.3	&78.3	&78.7 	&73.1 	&78.4 
 \\
        $\clubsuit$MulT &\textcolor{red!45}{71.4} 	&\textcolor{red!45}{60.3} 	&\textcolor{red!45}{66.4} 	&\textcolor{red!70}{78.6} 	&50.0 	&69.1 	&90.5 	&90.9 	&90.5 	&86.1	&50	&79.6	&\textcolor{red!45}{75.0}	&\textcolor{red!45}{74.6}	&\textcolor{red!45}{75.0}	&\textcolor{red!100}{100}$^{\uparrow}$	&\textcolor{red!100}{100}$^{\uparrow}$	&\textcolor{red!100}{100}$^{\uparrow}$	&\textcolor{red!70}{83.6} 	&\textcolor{red!70}{71.0} 	&\textcolor{red!70}{80.1} 
 \\
        $\spadesuit$MMIM &65.7 	&50.0 	&52.1 	&\textcolor{red!70}{78.6} 	&50.0 	&69.1 	&91.5 	&91.8 	&91.6 	&\textcolor{red!100}{87.3}	&54.5	&82.5	&74.0	&73.1	&73.8	&72.2	&79.2	&72.7	&78.2 	&66.4 	&73.6 
 \\
        $\spadesuit$LMF &65.7 	&50.0 	&52.1 	&\textcolor{red!70}{78.6} 	&53.0 	&72.0 	&\textcolor{red!70}{92.1} 	&\textcolor{red!70}{92.4} 	&\textcolor{red!70}{92.1} 	&86.1	&50	&79.6	&\textcolor{red!100}{78.1}	&\textcolor{red!100}{76.3}	&\textcolor{red!100}{77.5}	&\textcolor{red!100}{100}$^{\uparrow}$	&\textcolor{red!100}{100}$^{\uparrow}$	&\textcolor{red!100}{100}$^{\uparrow}$	&\textcolor{red!45}{83.4} 	&\textcolor{red!45}{70.3} 	&\textcolor{red!45}{78.9} 
 \\
        $\spadesuit$TFN &65.7 	&50.0 	&52.1 	&76.8 	&\textcolor{red!70}{58.0} 	&\textcolor{red!70}{74.4} 	&91.5 	&91.9 	&91.6 	&\textcolor{red!100}{87.3}	&54.5	&82.5	&72.9	&72.2	&72.8	&88.9	&91.7	&89.2	&80.5 	&69.7 	&77.1 
 \\
        $\spadesuit$Attention &\textcolor{red!100}{74.3} 	&\textcolor{red!70}{62.5} 	&\textcolor{red!70}{68.7} 	&76.8 	&\textcolor{red!70}{58.0} 	&\textcolor{red!70}{74.4} 	&\textcolor{red!100}{93.1}$^{\uparrow}$ 	&\textcolor{red!100}{93.4}$^{\uparrow}$ 	&\textcolor{red!100}{93.1}$^{\uparrow}$ 	&\textcolor{red!100}{87.3}	&\textcolor{red!70}{62.2}	&\textcolor{red!70}{85.2}	&\textcolor{red!70}{77.1}	&\textcolor{red!70}{75.1}	&\textcolor{red!70}{76.3}	&\textcolor{red!70}{94.4}	&\textcolor{red!70}{95.8}	&\textcolor{red!70}{94.5}	&\textcolor{red!100}{83.8} 	&\textcolor{red!100}{74.5} 	&\textcolor{red!100}{82.0} 
 \\
      \bottomrule
    \end{tabular}
    }
\end{threeparttable}
\end{table*}

\subsubsection{ASD Detection Results}
In the ASD detection task, the speech modality substantially outperforms the text modality. WavLM-Large again delivers the best performance in the speech modality. As shown in Table \ref{tab:monomodal_asd}, it achieves WF1 scores of 95.9\% and 81.5\% on the CPSD and ASDBank datasets, respectively. Its strong performance is attributed to its large-scale pre-training and denoising mechanisms. The text modality's performance in ASD detection is limited by dataset characteristics. 
For example, text classification is less meaningful for the French-language CPSD dataset, as many samples have identical text transcripts despite different labels. On the Dutch-language ASDBank dataset, large language models, such as the multilingual Qwen2.5-7B and Chinese-centric Baichuan2-7B, perform well, achieving WF1 scores of 76.3\% and 73.7\%, respectively. Notably, the performance of English-centric models is generally poor on this dataset. In contrast, the multilingual mGTE-Base underperforms compared to other multilingual LLMs, achieving a WF1 of only 63.2\%, which may be due to the limited proportion of Dutch data in its pre-training corpora.

Overall, the speech modality substantially outperforms the text modality in ASD detection, with average WF1 scores of 77.5\% and 65.3\%, respectively. This highlights the importance of acoustic features in capturing ASD-related patterns.

\begin{table}[!t]
  \centering
  \caption{Mono-modal and multi-modal benchmark for ASD detection. In this table, we report mono-modal and multi-modal results for all ASD datasets under the same experimental setup. 
  The definitions of all markers are the same as in Table~\ref{tab:monomodal_AD}.}
  \label{tab:monomodal_asd}
  \begin{threeparttable}
  \resizebox{.5\textwidth}{!}{
      \begin{tabular}{cccccccccc}
        \toprule
         Model  &\multicolumn{3}{c}{CPSD} &\multicolumn{3}{c}{ASDBank} &\multicolumn{3}{c}{Mean}\\
                $[\%]$ &WA	&UA &WF1 &WA &UA &WF1 &WA &UA  &WF1   \\
         \midrule
         \multicolumn{10}{c}{Acoustic} \\
         \midrule
        Wav2vec 2.0 B  &84.0 	&78.3 	&83.1 	&57.9 	&55.6 	&47.1 	&71.0 	&67.0 	&65.1 
\\
        Wav2vec 2.0 L &84.7 	&81.0 	&84.4 	&57.9 	&55.8 	&49.9 	&71.3 	&68.4 	&67.2 
\\
        XLS-R 300M  &92.9 	&90.1 	&92.7 	&63.2 	&63.1 	&63.2 	&78.1 	&76.6 	&78.0 
\\
        HuBERT B  &91.3 	&88.2 	&91.1 	&73.7	&73.9 	&73.7 	&\textcolor{red!45}{82.5} 	&81.1 	&\textcolor{red!45}{82.4} 
 \\
        HuBERT L &91.7 	&89.4 	&91.6 	&\textcolor{red!70}{78.9} 	&\textcolor{red!70}{78.6} 	&\textcolor{red!70}{78.8} 	&\textcolor{red!70}{85.3} 	&\textcolor{red!70}{84.0} 	&\textcolor{red!70}{85.2} 
  \\
        ExHuBERT  &85.8 	&82.1 	&85.5 	&71.1 	&70.8 	&71.0 	&78.5 	&76.5 	&78.3 
  \\
        Data2vec B &81.9 	&76.4 	&81.1 	&68.4 	&66.9 	&65.5 	&75.2 	&71.7 	&73.3 
 \\
        Data2vec L &84.9 	&81.4 	&84.6 	&68.4 	&68.6 	&68.4 	&76.7 	&75.0 	&76.5 
\\
        WavLM B   &91.2 	&89.2 	&91.1 	&73.7 	&73.9 	&73.7 	&\textcolor{red!45}{82.5} 	&\textcolor{red!45}{81.6} 	&\textcolor{red!45}{82.4} 
\\
        \underline{WavLM L}  &\textcolor{red!100}{96.0} 	&\textcolor{red!100}{94.2} 	&\textcolor{red!100}{95.9} 	&\textcolor{red!100}{81.6} 	&\textcolor{red!100}{81.9} 	&\textcolor{red!100}{81.5} 	&\textcolor{red!100}{88.8} 	&\textcolor{red!100}{88.1} 	&\textcolor{red!100}{88.7} 
 \\
        Whisper S &84.0 	&78.4 	&83.1 	&\textcolor{red!70}{76.3} 	&\textcolor{red!45}{75.6} 	&\textcolor{red!70}{75.7} 	&80.2 	&77.0 	&79.4 
   \\
        Whisper M  &84.5 	&79.8 	&83.9 	&68.4 	&68.3 	&68.4 	&76.5 	&74.1 	&76.2 
 \\
        Emotion2vec  &\textcolor{red!70}{94.9} 	&\textcolor{red!70}{93.3} 	&\textcolor{red!70}{94.8} 	&60.5 	&60.0 	&60.1 	&77.7 	&76.7 	&77.5 
  \\
        Wav2vec 2.0 CN  &92.3 	&90.3 	&92.2 	&55.3 	&56.1 	&54.2 	&73.8 	&73.2 	&73.2 
 \\
        HuBERT CN   &\textcolor{red!45}{93.0} 	&\textcolor{red!45}{90.4} 	&\textcolor{red!45}{92.9} 	&65.8 	&65.8 	&65.8 	&79.4 	&78.1 	&79.4  
   \\
        \midrule
         \multicolumn{10}{c}{Linguistic} \\
         \midrule
        RoBERTa L &-$^*$	&-	&-	&\textcolor{red!70}{73.7} 	&\textcolor{red!70}{73.6} 	&\textcolor{red!70}{73.7} 	&\textcolor{red!70}{73.7} 	&\textcolor{red!70}{73.6} 	&\textcolor{red!70}{73.7} 

 \\ 
        DeBERTa L &-	&-	&-	&60.5 	&60.3 	&60.4 	&60.5 	&60.3 	&60.4 
 \\ 
        XLNet L   &-	&-	&-	&60.5  &61.1 &60.1  &60.5  &61.1 &60.1
   \\ 
        SimCSE-     \\
        RoBERTa L   &-	&-	&-	&57.9  &57.2	&57.2 	&57.9  &57.2	&57.2
\\ 
        E5 B   &-	&-	&-	&63.2  &62.2	&61.9 	&63.2  &62.2	&61.9
\\
        \underline{E5 L}  &-	&-	&-	&63.2 &62.5	&62.5 	&63.2 &62.5	&62.5
  \\
        GTE B  &-	&-	&-	&63.2  &63.6	&63.0 	&63.2  &63.6	&63.0
 \\
        GTE L    &-	&-	&-	&63.2 &62.8	&63.0 	&63.2 &62.8	&63.0
   \\
        mGTE B   &-	&-	&-	&63.2 &63.1	&63.2 	&63.2 &63.1	&63.2
 \\
        mE5 L   &-	&-	&-	&63.2  &63.3	&63.2 	&63.2  &63.3	&63.2
\\
        Llama-3.1-8B   &-	&-	&-	&71.1 &70.8	&71.0 &71.1 &70.8 &71.0
 \\
        Baichuan2-7B  &-	&-	&-	&\textcolor{red!70}{73.7} &\textcolor{red!45}{73.3}	&\textcolor{red!45}{73.5} 	&\textcolor{red!70}{73.7} &\textcolor{red!45}{73.3}	&\textcolor{red!45}{73.5}
 \\
        Qwen2.5-7B   &-	&-	&-	&\textcolor{red!100}{76.3} &\textcolor{red!100}{76.4}	&\textcolor{red!100}{76.3} 	&\textcolor{red!100}{76.3} &\textcolor{red!100}{76.4}	&\textcolor{red!100}{76.3}
 \\

          \midrule
         \multicolumn{10}{c}{Multi-Model Fusion} \\
         \midrule
        $\clubsuit$MFM &66.2 	&50.0 	&52.7 	&60.5 	&59.7 	&59.5 	&63.4 	&54.9 	&56.1 
\\
        $\clubsuit$MFN &92.8 	&90.6 	&92.7 	&63.2 	&62.5 	&62.5 	&78.0 	&76.6 	&77.6 
 \\
        $\clubsuit$GraphMFN  &83.4 	&78.9 	&82.8 	&68.4 	&68.1 	&68.2 	&75.9 	&73.5 	&75.5 
 \\
        $\clubsuit$MulT &\textcolor{red!70}{94.0} 	&\textcolor{red!70}{91.8} 	&\textcolor{red!70}{93.9} 	&\textcolor{red!100}{78.9} 	&\textcolor{red!100}{79.2} 	&\textcolor{red!100}{78.9} 	&\textcolor{red!100}{86.5} 	&\textcolor{red!100}{85.5} 	&\textcolor{red!100}{86.4} 
 \\
        $\spadesuit$MMIM &91.9 	&90.8 	&91.9 	&\textcolor{red!70}{73.7} 	&\textcolor{red!70}{72.8} 	&\textcolor{red!70}{72.8} 	&\textcolor{red!70}{82.8} 	&\textcolor{red!70}{81.8} 	&\textcolor{red!70}{82.4} 
 \\
        $\spadesuit$LMF &92.9 	&90.1 	&92.7 	&\textcolor{red!45}{71.1} 	&\textcolor{red!45}{71.4} 	&\textcolor{red!45}{71.0} 	&82.0 	&80.8 	&81.9 
 \\
        $\spadesuit$TFN &\textcolor{red!100}{94.3} 	&\textcolor{red!100}{92.5} 	&\textcolor{red!100}{94.2} 	&63.2 	&61.9 	&61.0 	&78.8 	&77.2 	&77.6 
 \\
        $\spadesuit$Attention  &\textcolor{red!70}{94.0} 	&\textcolor{red!70}{91.8} 	&\textcolor{red!70}{93.9} 	&71.1 	&70.8 	&71.0 	&\textcolor{red!45}{82.6} 	&\textcolor{red!45}{81.3} 	&\textcolor{red!45}{82.5} 
 \\
      \bottomrule
    \end{tabular}
}
    \begin{tablenotes}  
      \item[*] no results due to its lack of text modality.
    \end{tablenotes}
\vspace{-.3cm}
\end{threeparttable}
\end{table}

\subsubsection{Key Findings}  
Based on the mono-modal results across the three diseases, we summarize the following conclusions:  
(1) WavLM-Large excels in speech tasks: Its superior performance in AD and ASD detection demonstrates the importance of large-scale pre-training and advanced denoising mechanisms.  
(2) Emotion-related features are crucial: Emotion2vec-Base performs well across all three tasks, highlighting the significance of emotion-related paralinguistic information.  
(3) Large models generally outperform base models: Across all tasks and modalities, large-scale models (\eg WavLM-Large, E5-Large) consistently achieve better performance than their base counterparts, emphasizing the importance of model scale in capturing complex disease patterns.  
(4) Language capability is critical for text models: Our multi-lingual evaluation consistently shows that the performance of text-based models is heavily dependent on their language coverage. English-centric models struggle on non-English datasets, while multilingual models (e.g., mE5-L, mGTE-Base) demonstrate greater robustness, confirming the necessity of matching model capability with task language.
(5) Modality effectiveness is disease-specific: Text features are more effective for AD detection, speech and text perform comparably for depression detection, and speech features are more effective for ASD detection, indicating disease-specific modality correlations.  

\vspace{-.3cm}
\subsection{Multi-Modal Benchmark}
This section analyzes the multi-modal experimental results of the FEND framework for detecting AD, depression, and ASD. We evaluate the performance of multi-modal fusion by combining the best-performing mono-modal models across the three tasks: \textit{WavLM-large} for speech and \textit{E5-large} for text, which were identified as the top-performing models in their respective modalities based on their overall performance across all diseases. The results are summarized in Tables \ref{tab:monomodal_AD}, \ref{tab:monomodal_depression}, and \ref{tab:monomodal_asd}.

\subsubsection{AD Detection Results}
In the AD detection task, multi-modal fusion improves performance on most monolingual datasets. For example, on the ADReSS dataset, the combination of WavLM-Large and E5-Large achieves a WF1 of 85.4\%, outperforming the best mono-modal model (E5-Large with 83.3\% WF1) (see Table \ref{tab:monomodal_AD}). Similarly, on the PITT dataset, the multi-modal approach achieves a WF1 of 84.3\%, compared to 83.5\% for the best mono-modal model. However, on the cross-lingual ADReSS-M dataset, multi-modal fusion underperforms. This is primarily due to the severe challenge posed by the language mismatch (English to Greek), which degrades the quality of representations from both modalities. The scarcity of target-language (Greek) samples in the training set makes effective learning for the test set nearly impossible, an issue distinct from the overall training data volume.

\subsubsection{Depression Detection Results}
In depression detection, multi-modal fusion enhances performance on larger datasets. As shown in Table \ref{tab:monomodal_depression}, Attention achieves a WF1 of 93.1\% on the D-VLOG dataset, surpassing the best mono-modal model (E5-Large with 92.6\%), while on the E-DAIC dataset, the multi-modal approach reaches 78.6\% versus 75.7\% for the best mono-modal model. However, on smaller datasets (\eg EATD, MODMA), mono-modal models perform better, possibly due to overfitting in multi-modal fusion.

\subsubsection{ASD Detection Results}
In the ASD detection task, multi-modal fusion fails to surpass the mono-modal baselines on both the CPSD and ASDBank datasets. As shown in Table \ref{tab:monomodal_asd}, TFN achieves a WF1 of 94.2\% on the CPSD dataset, which is comparable to the best mono-modal model (WavLM-Large with 95.9\% WF1). On the ASDBank dataset, the  multi-modal approach achieves a WF1 of 78.9\%, underperforming compared to the best mono-modal model (WavLM-large with 81.5\% WF1).

\subsubsection{Key Findings}  
Based on the multi-modal fusion results, we summarize the following conclusions:  
(1) Multi-modal fusion is effective for AD and depression detection: Combining speech and text information generally improves performance, especially with attention-based and tensor fusion methods.  
(2) ASD detection remains challenging: Multi-modal fusion fails to surpass mono-modal baselines, likely due to dataset heterogeneity.  
(3) Attention mechanisms are robust: Attention-based methods (\eg Attention, MulT) perform well across tasks and datasets, demonstrating their ability to capture interactions between modalities.  
(4) Modality imbalance affects performance: In some datasets, multi-modal fusion does not outperform the best mono-modal model, likely due to disparities in information quality and contribution between speech and text modalities, particularly in cross-lingual tasks.  

\begin{figure}[!t]
  \centering
  \includegraphics[width=\linewidth]{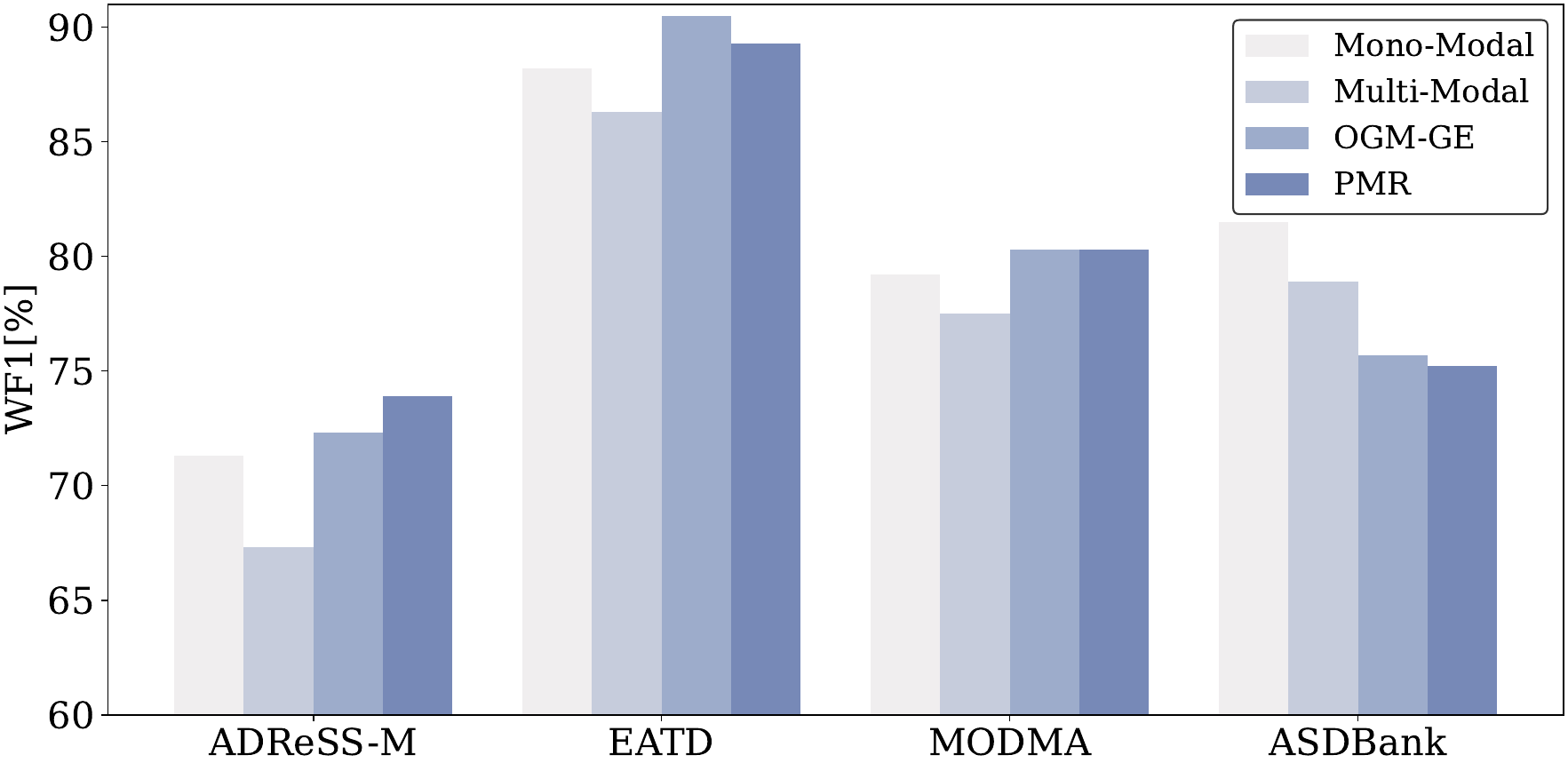}
  \caption{Performance comparison of multi-modal fusion with and without modality imbalance mitigation techniques (OGM-GE and PMR) across different datasets. Mono-modal: The best mono-modal performance. Multi-modal: The best multi-modal performance.}
  \label{modal_imbalance}
  \vspace{-.2cm}
\end{figure}

\vspace{-.3cm}
\subsection{Modality Imbalance Analysis}
Multi-modal fusion fails to outperform the best mono-modal models on datasets like ADReSS-M, EATD, MODMA, and ASDBank. These datasets are either multi-lingual or non-English, and their underperformance may be attributed to modality imbalance, where multi-modal models optimize for the dominant modality while neglecting others, leading to suboptimal performance.

To address this, we employ HAFFormer as the encoder and utilize Concat and Sum as fusion algorithms, optimized with On-the-fly Gradient Modulation-Generalization Enhancement (OGM-GE) \cite{peng2022balanced} and Prototypical Modal Rebalance (PMR) \cite{fan2023pmr}, which are specifically designed to mitigate modality imbalance. Results in Fig~\ref{modal_imbalance} show that OGM-GE and PMR improve performance on most datasets, surpassing the best mono-modal models, demonstrating the effectiveness of addressing modality imbalance in multi-modal fusion.

\begin{figure*}[!t]
  \centering
  \includegraphics[width=.9
  \linewidth]{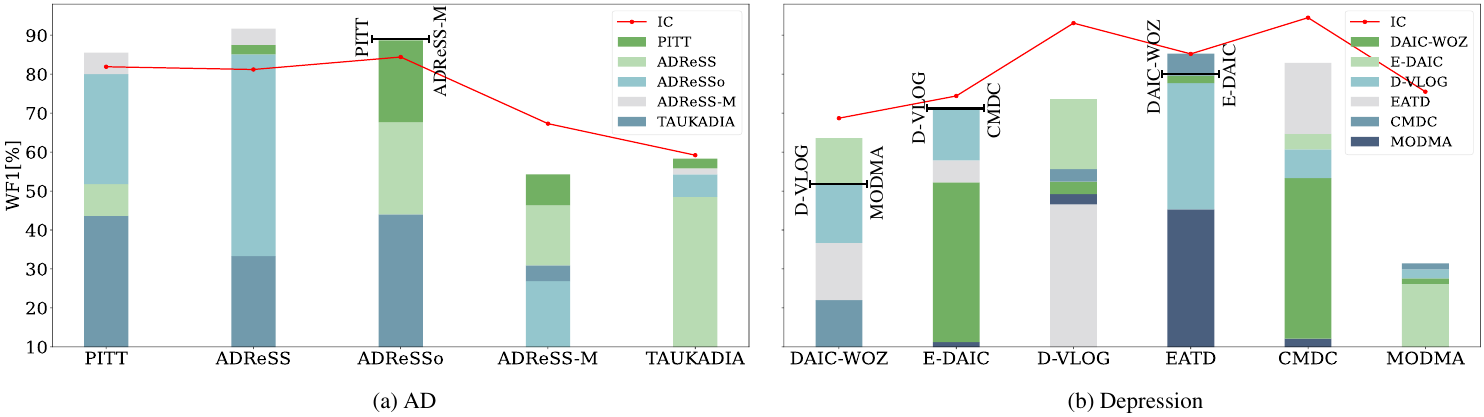}
  \caption{Cross-corpus inference results. (a) and (b) present the cross-corpus inference results for AD and depression, respectively. The x-axis denotes the test datasets. Legends represent the WF1 scores of models pre-trained on different datasets when evaluated on the test datasets.
  ``IC'' denotes the ``Intra-Corpus'' experimental results. 
  The ``\textbf{---}'' indicates the same performance obtained on the two marked datasets. }
  \label{cross-corpus}
  \vspace{-.4cm}
\end{figure*}

\vspace{-.3cm}
\subsection{Cross-Corpus Analysis}
To evaluate the generalization ability and robustness of the FEND framework, we conducted cross-corpus experiments, where models were trained on one dataset and tested on another. For each disease, the best-performing multi-modal fusion algorithm was used: MMIM for AD and ASD, and Attention for depression detection. The experimental results, shown in Fig~\ref{cross-corpus}, reveal several key observations. First, models trained on fully English datasets (\eg ADReSS, PITT, DAIC-WOZ) exhibit strong generalization, with WF1 scores mostly exceeding 70\%-80\%, which can be attributed to consistent task objectives and language characteristics. Second, models trained on multi-lingual datasets (\eg ADReSS-M) perform poorly on datasets with language inconsistencies, as differences in vocabulary, grammar, and expression hinder effective feature transfer. Third, models trained on datasets with unique tasks or limited sizes, such as MODMA, show poor generalization across all datasets due to task heterogeneity and insufficient training data. 

Additionally, the cross-corpus generalization of ASD detection is notably poor, primarily due to substantial differences in task design and language characteristics between datasets like CPSD (French) and ASDBank (Dutch). Hence, the cross-corpus experimental results for ASD are not included in Fig~\ref{cross-corpus}.

\vspace{-.3cm}
\subsection{Discussions}
Our experiments highlight the potential of multi-modal fusion for neuropsychiatric disease detection while revealing its limitations. Below, we discuss the key implications of our findings:

\subsubsection{Effectiveness of Multi-Modal Fusion}
Multi-modal fusion demonstrates considerable improvements in depression and AD detection, particularly with Attention-based and tensor fusion methods. For example, the Attention model achieves a WF1 of 93.1\% on the D-VLOG dataset, outperforming the best mono-modal model. However, ASD detection remains challenging, likely due to dataset heterogeneity and modality imbalance, where speech contains more discriminative information than text. This highlights the need for tailored fusion strategies for different diseases.

\subsubsection{Challenges of Multi-Modal Fusion Methods}
Current multi-modal fusion methods still face challenges in neuropsychiatric disease detection. While they show effectiveness on some datasets, not all fusion methods yield optimal results. Notably, these methods are primarily derived from emotion recognition or sentence classification domains and often designed for three-modality fusion, which may not fully align with the characteristics of neuropsychiatric diseases. For instance, the modality interaction mechanisms in emotion recognition may fail to capture the unique multi-modal patterns of neuropsychiatric disorders. Therefore, there is a pressing need to develop and open-source multi-modal fusion methods specifically tailored for neuropsychiatric disease detection to enhance model adaptability and performance.

\subsubsection{Robustness of Attention Mechanisms}
Attention-based methods (\eg Attention, MulT) exhibit strong performance across tasks and datasets, demonstrating their ability to model complex interactions between modalities. For instance, MulT achieves the highest average WF1 across all six depression datasets. However, attention mechanisms are not universally optimal; for example, in AD detection, tensor fusion methods (\eg TFN) outperform attention-based models. This suggests that the choice of fusion strategy should be task-specific.

\subsubsection{Modality Imbalance}
Modality imbalance significantly impacts multi-modal fusion performance, particularly in multi-lingual tasks. For instance, in ASD detection, the speech modality dominates due to its richer information content, making it difficult for models to effectively utilize the text modality. Our experiments with OGM-GE and PMR demonstrate that these methods can mitigate modality imbalance to some extent, improving performance in certain scenarios. However, they cannot fully resolve the issue, highlighting the need for further research to develop more robust solutions for diverse datasets and tasks. A root cause of this imbalance in multi-lingual scenarios is often the limited language capability of the text-based foundation models. Our analysis consistently shows that English-centric models struggle to generate meaningful representations for non-English languages, leading to a low-quality text modality that is subsequently outweighed by the more robust speech modality. This underscores that effective multi-modal fusion in diverse linguistic contexts is contingent not only on the fusion algorithm itself, but on the underlying unimodal models possessing adequate language coverage.

\subsubsection{Disease-Specific Modality Correlations}
Our results reveal disease-specific correlations between modalities. Text features are more effective for AD detection, likely due to the linguistic abnormalities associated with cognitive decline. In contrast, speech features dominate in ASD detection, as acoustic patterns better capture behavioral abnormalities. Depression detection shows a balance between speech and text modalities, suggesting that both acoustic and linguistic cues are important for emotional regulation. These findings emphasize the need for disease-specific multi-modal fusion strategies.

\subsubsection{Generalization Challenges and Real-World Implications}
Furthermore, our cross-corpus analysis highlights significant generalization challenges, which have profound implications for real-world deployment. The observed performance drop when models are tested on datasets with different languages, recording conditions, or subtle task variations underscores their sensitivity to domain shifts. This suggests that deploying a model trained on a single academic dataset directly into a diverse clinical environment is likely to fail without proper adaptation. Consequently, a critical direction for future work is the development of robust models that excel in domain adaptation and generalization. Strategies such as multi-source training, unsupervised domain adaptation, and the design of features invariant to nuisance variables are essential to bridge the gap between research-setting performance and practical clinical utility.

\subsubsection{Impact of ASR Quality}
It is important to acknowledge the potential impact of ASR quality on our findings for the text modality. While we employed powerful ASR systems (SenseVoice for Chinese and Whisper-Large for other languages), transcription errors are a potential source of noise. Such errors could affect the quality of text representations extracted by foundation models, thereby influencing the final detection performance. One of our goals was to assess model performance under practical conditions where ASR-generated transcripts are standard. Nevertheless, developing models that are more robust to ASR errors or leveraging ASR confidence scores represents a significant avenue for future research to further enhance the reliability of text-based analysis.

\section{CONCLUSION AND FUTURE WORKS} \label{sec:conclusion}

This paper introduces FEND, a unified evaluation framework that addresses the critical need for standardized, multi-dimensional benchmarking of foundation models in neuropsychiatric disorder detection. Through an extensive empirical study, we have demonstrated the strengths and weaknesses of various acoustic and linguistic models, highlighting the potential of multi-modal fusion for AD and depression while also revealing significant challenges, such as modality imbalance and limited cross-lingual generalization. Our findings provide the community with valuable benchmarks and deep theoretical insights, intended to guide future model selection, methodology design, and the overall trajectory of automated neuropsychiatric disease assessment.

In the future, we plan to extend this research in the following directions: (1) Developing specialized multi-modal fusion methods tailored to neuropsychiatric disease detection, moving beyond existing approaches adapted from emotion recognition or sentence classification domains; (2) Exploring advanced multi-modal fusion techniques, such as adaptive weight adjustment and cross-modal learning strategies, as well as domain adaptation methods to address the specific cross-lingual challenges identified in datasets like ADReSS-M; (3) Creating domain-specific foundation models optimized for neuropsychiatric disease detection, incorporating domain-specific knowledge and refined pre-training datasets; 
(4) Mitigating modality imbalance through adaptive fusion techniques that dynamically adjust modality contributions based on their relevance and quality; 
(5) Validating the framework on larger and more diverse datasets, including additional modalities such as physiological signals and imaging data, to build more comprehensive diagnostic models;
(6) Extending the FEND framework to benchmark emerging model architectures, including end-to-end multi-modal models like AudioLLMs, as well as LLMs specifically fine-tuned for high-quality text embedding generation;
and (7) Investigating advanced training paradigms, such as full end-to-end fine-tuning or parameter-efficient fine-tuning (PEFT), for the most promising model combinations identified in this work. This will allow us to explore the upper bounds of performance and better adapt the foundation models to the specific nuances of neuropsychiatric disorder detection.




\bibliographystyle{IEEEtran}
\bibliography{IEEEabrv,dzr}

\vfill

\end{document}